\documentclass[]{fairmeta}

\title{Learning Graph Quantized Tokenizers}

\author[\dagger]{Limei Wang}
\author[\dagger]{Kaveh Hassani}
\author[]{Si Zhang}
\author[]{Dongqi Fu}
\author[]{Baichuan Yuan}
\author[]{Weilin Cong}
\author[]{Zhigang Hua}
\author[]{Hao Wu}
\author[]{Ning Yao}
\author[]{Bo Long}

\affiliation[]{Meta AI}
\contribution[\dagger]{Equal Contributions}

\abstract{Transformers serve as the backbone architectures of Foundational Models, where domain-specific tokenizers allow them to adapt to various domains. Graph Transformers (GTs) have recently emerged as leading models in geometric deep learning, outperforming Graph Neural Networks (GNNs) in various graph learning tasks. However, the development of tokenizers for graphs has lagged behind other modalities. To address this, we introduce GQT (\textbf{G}raph \textbf{Q}uantized \textbf{T}okenizer), which decouples tokenizer training from Transformer training by leveraging multi-task graph self-supervised learning, yielding robust and generalizable graph tokens. Furthermore, the GQT utilizes Residual Vector Quantization (RVQ) to learn hierarchical discrete tokens, resulting in significantly reduced memory requirements and improved generalization capabilities. By combining the GQT with token modulation, a Transformer encoder achieves state-of-the-art performance on 20 out of 22 benchmarks, including large-scale homophilic and heterophilic datasets.
}

\metadata[Code]{\url{https://github.com/limei0307/GQT}}

\begin{document}

\maketitle

\section{Introduction} \label{sec:introduction}
Following the success of Transformers \citep{NIPS2017_3f5ee243} in Natural Language Processing \citep{devlin-etal-2019-bert, NEURIPS2020_1457c0d6} and Computer Vision  \citep{dosovitskiy2021an}, Graph Transformers (GTs) \citep{dwivedi2020generalization, ying2021transformers, rampavsek2022recipe, pmlr-v202-shirzad23a, chen2023nagphormer, wu2022nodeformer} have emerged as strong models in geometric deep learning.
In contrast to message-passing Graph Neural Networks (GNNs), which are inherently constrained by strong locality inductive biases \citep{battaglia2018relationalinductivebiasesdeep, veličković2018graph, Hou2020Measuring, hamilton2017inductive, kipf2017semisupervised, DBLP:conf/TMLR/ZhengFMH24}, GTs exhibit greater expressivity due to their capacity to capture long-range interactions between nodes \citep{pmlr-v202-ma23c,kim2022pure, zopf20221}. This is particularly beneficial in heterophilic settings where local alignment does not hold \citep{fu2024vcrgraphormer}.
This dichotomy highlights a fundamental trade-off between GNNs, which focus on local neighborhood aggregation, and GTs, which employ pairwise attention to model global graph structures. A natural question arises: can we synergistically integrate the strengths of both approaches to leverage the complementary benefits of local and global representations? Specifically, is it possible to harness the locality-aware representations learned by GNNs to construct discrete tokens, thereby enabling GTs to operate efficiently while still capturing salient graph properties?

GTs require consideration of both graph structure and features, as nodes with identical features will otherwise be projected into the same representation regardless of their surrounding structures \citep{hoang2024survey}. There are three general approaches to address this limitation \citep{hoang2024survey}: (1) node feature modulation, which involves injecting  structural information into the node features; (2) context node sampling, where a sampling strategy is used to construct a sequence over the neighbor nodes; and (3) modifying the  architecture of a vanilla Transformer to directly incorporate structural biases.
Given that Transformers are universal approximators of sequence-to-sequence functions \citep{Yun2020Are} and considering the rapid developments in efficient implementation of Multi-Head Attention (MHA) module \citep{dao2022flashattention, liu2024ringattention}, which enables longer context sizes \citep{reid2024gemini}, we argue that a well-designed graph tokenizer can allow a vanilla Transformer to efficiently process even large-scale graphs.
Recent studies on applying Large Language Models (LLMs) to Text-Attributed Graphs (TAGs) have shown surprisingly strong performance gains surpassing those of GNNs, suggesting that vanilla Transformers are indeed capable of effectively learning graph structures \citep{ye-etal-2024-language, xu2025makellmsstrongnode}. Nonetheless, LLMs are not efficient at inference time. Our goal is to devise a lightweight and efficient graph tokenizater that allows vanilla Transformer encoders to learn graph structures effectively.

Tokenizers typically employ self-supervised objectives to abstract data into a sequence of discrete tokens, allowing Transformers to learn representations across various modalities as a unified stream of data. The discretization is usually achieved through vector quantization techniques \citep{van2017neural, lee2022autoregressive}, which offer several benefits, including: (1)  significantly reduced memory requirements, (2) improved inference efficiency, (3) allowing Transformers to focus on long-range dependencies rather than local information, and (4) the capacity to learn more high-level representations due to a compact latent space  \citep{Yuan_2021_ICCV, yu2022vectorquantized}. These advantages are particularly important in auto-regressive generative modeling, where quantized tokens allow Transformers to generate high-quality outputs in multiple modalities \citep{dubey2024llama, lee2022autoregressive, dhariwal2020jukebox, pmlr-v139-ramesh21a, team2024chameleon}. Despite its importance in other domains, tokenization remains under-explored for graph-structured data. To address this limitation, we propose the \textbf{Graph Quantized Tokenizer (GQT)}, a novel approach that learns a hierarchical sequence of tokens over graphs using self-supervised objectives tailored to graph-structured data. More specifically, our contributions are as follows:
\begin{itemize}
    \item We propose a graph tokenizer that uses multi-task graph self-supervised learning to train a graph encoder, enabling it to fully capture local interactions and allowing the Transformer to focus on long-range dependencies.
    \item Our approach adapts Residual Vector Quantization (RVQ) within the graph  tokenizer to learn hierarchical discrete tokens, resulting in significantly reduced memory requirements and improved generalization capabilities.
    \item We introduce a novel combination of semantic edges and random walks to facilitate access to long-range interactions, and employ hierarchical encoding and gating mechanisms to modulate the tokens and provide informative representations to the Transformer.
    \item Through extensive experiments on both homophilic and heterophilic datasets, including large-scale and long-range benchmarks, we demonstrate that our tokenizer enables Transformer encoders to achieve state-of-the-art performance on 20 out of 22 benchmarks while substantially reducing the memory footprint of the embeddings.
\end{itemize}

\section{Related Works} \label{sec:related_works}

\textbf{Graph Transformers (GTs)} have shown promising performance on various graph learning tasks, surpassing GNNs on many benchmarks. Designing GTs can be broadly categorized into two directions \citep{hoang2024survey, muller2024attending}: (1) modifying the vanilla Transformer architecture to incorporate structural inductive biases, or (2) encoding the input graph to make it compatible with the vanilla Transformer. Early examples of the first approach include Graph Attention Network (GAT) \citep{veličković2018graph}, which uses an attention module to compute pairwise node attention and masks the attention matrix based on connectivity information. Subsequent works have replaced the scaled-dot attention module with various structure-aware sparse attention modules \citep{rampavsek2022recipe, bo2023specformer, ying2021transformers, deng2024polynormer, wu2023simplifying, 10.24963/ijcai.2023/244, pmlr-v162-chen22r, dwivedi2020generalization, pmlr-v202-shirzad23a, pmlr-v202-ma23c}. Graph Memory Network (GMN) \citep{Khasahmadi2020Memory-Based} is an example of the second approach, which passes non-linear projections of node features  and structural encoding to a Transformer-like model. Structural encodings such as Laplacian eigenvectors or Random walk-based encoding \citep{dwivedi2022graph, pmlr-v202-ma23c, pmlr-v235-canturk24a}, allow injecting structural information directly into the node features. Some works use GNNs to encode local structure along with node features into embeddings that are passed to vanilla Transformers to capture long-range dependencies \citep{rong2020self, wu2021representing, chen2023nagphormer, pmlr-v162-chen22r}. Recent  studies leverage LLMs, where graphs are represented through natural language, and an LLM performs graph-related tasks through in-context learning, instruction-tuning, or soft-prompting \citep{fatemi2024talk, ye-etal-2024-language, he2024harnessing}. For a detailed survey on GTs, see \citep{muller2024attending, hoang2024survey}.

\textbf{Graph Tokenization} provides GTs with rich node tokens that encapsulate both structural and semantic information. TokenGT \citep{kim2022pure} treats nodes and edges as independent  tokens defined by their features, type identifiers, and structural encodings. NAGphormer \citep{chen2023nagphormer} represents each node with $L$ tokens, where the $l^{th}$ token is the  representation of the node from the $l^{th}$ hop aggregation. GraphiT \citep{mialon2021graphit} defines a node token as the concatenation of its feature and representation from a  graph convolutional kernel network (GCKN). VCR-Graphormer \citep{fu2024vcrgraphormer} expands the notion of node tokens to include sequences comprising the node feature and features of  semantically and community-related neighboring nodes. SGT \citep{liu2023rethinking} is a non-parametric tokenizer designed for molecular tasks, which unlike motif-based tokenizers \citep{zhang2021motif, jin2018junction} or GNN pre-training methods \citep{xia2023molebert}, simplifies the tokenization process to a non-parametric graph operator without non-linearity.  NodePiece \citep{galkin2022nodepiece} is a knowledge-graph tokenizer that represents a target node as a hash of its top-k closest anchors, their distances, and relational context. While Vector Quantization (VQ) has been explored in other modalities \citep{van2017neural, lee2022autoregressive, yu2022point, van2024fast, li2024geometry}, its application in graph  learning is limited. Notable exceptions include VQ-GNN \citep{ding2021vqgnn}, which uses quantized representations combined with a low-rank graph convolution matrix to avoid neighbor explosion problem, VQGraph \citep{yang2024vqgraph}, which employs VQ for distilling a GNN into an MLP, and NID \citep{luo2024structure}, which uses VQ to learn discrete node IDs for downstream prediction tasks.

\section{Preliminaries} \label{sec:preliminaries}
\textbf{Messag-Passing GNNs}. Let $\mathcal{G}$ denote the space of graphs. A graph $g \in \mathcal{G}$ is defined as $\left(\mathcal{V}, \mathcal{E}, \textbf{X}, \textbf{E} \right)$ where $\mathcal{V}$ is the set of nodes and $\mathcal{E} \subseteq \mathcal{V} \times \mathcal{V}$ is the set of edges. $\textbf{X} \in \mathbb{R}^{\left|\mathcal{V}\right| \times d_x}$  represents the node features of dimension $d_x$, and $\textbf{E} \in \mathbb{R}^{\left|\mathcal{V}\right| \times \left|\mathcal{V}\right| \times d_e}$ represents the edge features of dimension  $d_e$. A message-passing GNN takes $g$ as input and learns representations $h^l_v$ for $v \in \mathcal{V}$ ($h^0_v=x_v$) in each layer $l$ as follows \citep{gilmer2017neural}:
\begin{equation}
    h^l_v = f^l_{\theta}\left(h^{l-1}_v, g^l_{\phi} \left(\left\{\left(h^{l-1}_v, h^{l-1}_u, e_{uv} \right) | u \in \mathcal{N}_v\right\}\right) \right)
\end{equation}
where $f_{\theta}$ and $g_{\phi}$ are known as combine and aggregate functions, respectively. $\mathcal{N}_v$ denotes the set of immediate neighbors of the node $v$. Once the node representations are computed, we can perform various tasks including node classification as $\text{MLP}\left(h_v\right)$, edge prediction as $\text{MLP}\left(h_u \odot h_v\right)$, or graph classification as $\text{MLP}\left(\mathcal{R}\left(\left\{h_u |u \in \mathcal{V}\right\}\right)\right)$, where $\mathcal{R}$ is a pooling (readout) function.

\textbf{Graph Transformers} use a tokenizer $T_v = \mathcal{T}_{\psi}\left(\mathcal{N}(v)\right)$ to map each node $v \in \mathcal{V}$ into a sequence of tokens $T_v$ by considering a notion of neighborhood $\mathcal{N}$. The simplest design is when $\mathcal{N}$ is zero-hop neighborhood (i.e., the node itself) and $\mathcal{T}_{\psi}$ is a node  feature lookup function. The neighborhood $\mathcal{N}$ can be extended to include the node's ego network \citep{zhao2021gophormer} or top-k Random Walk based neighbors \citep{fu2024vcrgraphormer}, and $\mathcal{T}_{\psi}$ can be enhanced to representations from a GNN \citep{chen2023nagphormer}. Node tokens along with positional encodings ($\text{PE}$) are passed to the Transformer as $h^0_v = \left[T_v || \text{PE}\left(v\right) \right]$. The representations in the $l^{th}$ layer of a Transformer encoder are computed as:
\begin{align}
        h^l_v &= \text{LN}\left(\text{MHA}\left(\text{LN}\left(h^{l-1}_v\right)\right) + h^{l-1}_v\right) \\
        h^l_v &= h^l_v + \text{MLP}\left(h^l_v\right)
\end{align}
where LN and MHA are LayerNorm and multi-head attention, respectively. Similar to Transformer encoders in other modalities \citep{devlin-etal-2019-bert, dosovitskiy2021an},  we can append a special classification token ($\left[\text{CLS}\right]$) to the input and use its representation to perform various classification tasks on the graph: $\text{MLP}\left(h_{\left[\text{CLS}\right]}\right)$. 

\textbf{Vector Quantization} projects embeddings $\textbf{X} \in \mathbb{R}^{n \times d_x}$ into a more compact space of codebooks $\textbf{C} \in \mathbb{R}^{k \times d_c}$, where $k \ll n$. The  codebooks can be learned by minimizing various objectives such as K-means clustering. The new representation of $x_i$ is then computed as \citep{van2017neural}:
\begin{equation}
    z(x_i) = c_k \quad \text{where} \quad k=\argmin_j \Vert x_i - c_j \Vert_2^2
\end{equation}
Building upon this concept, Residual-VQ (RVQ) \citep{lee2022autoregressive} extends VQ to a sequence of codebooks, where each consecutive codebook quantizes the residual error from the previous codebook,  i.e., $r_i=z_i - c_k$. This hierarchical approach constructs a multi-level quantized representation, enhancing the overall quantization quality. More details of RVQ are included in Appendix~\ref{sec:appendix_model_details}.

\section{Self-Supervised Graph Tokenization} \label{sec:tokenizer}
\subsection{Tokenizer Properties}
Our goal is to design a graph tokenizer that learns node tokens that exhibit three key characteristics:

\textbf{Modeling Local Interactions}. The tokens should encapsulate local interactions, allowing the Transformer to focus on long-range dependencies. This is analogous to Vision Transformers (ViTs), where the Transformer attends to image patches instead of pixels \citep{dosovitskiy2021an, liu2021swin}. To achieve this, we leverage GNNs as the tokenizer encoder to model local interactions in the representation space \citep{battaglia2018relationalinductivebiasesdeep}. Our design accommodates various GNN layer choices without constraints; for simplicity, we opt for a GAT encoder \citep{veličković2018graph}. 

\textbf{Memory Efficiency}. The tokens also should be compact to facilitate efficient memory usage. To achieve this, we introduce a  Residual-VQ (RVQ) \citep{lee2022autoregressive} layer to quantize the GNN representations into a sequence of discrete tokens. Quantization not only helps with generalization due to its  regularization effect but also significantly reduces memory usage. Using an RVQ with $c$ codebooks (typically $c=\{2, \cdots, 8\}$), a graph with feature matrix  $\textbf{X} \in \mathbb{R}^{N \times d_x}$ can be represented as $\textbf{X}_Q \in  \mathbb{N}^{N \times c}$ and codebook representation of $\textbf{C} \in  \mathbb{R}^{c \times K \times d_c}$, where $c$ is the number of codebooks (i.e., levels of quantization), $K$ is the codebook size, and $d_c$ is the code dimension. To illustrate the benefits of this approach, consider a graph with $10^6$ nodes  and a feature dimension of 1024 ($\textbf{X} \in \mathbb{R}^{10^6 \times 1024}$). Using an RVQ with 3 codebooks and a codebook size of 256, this graph can be represented as $\textbf{X}_Q \in \mathbb{N}^{10^6 \times 3}$ plus $\textbf{C} \in  \mathbb{R}^{3 \times 256 \times 1024}$, resulting in a 270-fold reduction in memory. 

\textbf{Robustness and Generalization}. The tokens should be robust and generalizable. To achieve this, we rely on graph self-supervised learning. Self-supervised representations have been shown to be more robust to class imbalance \citep{liu2022selfsupervised} and distribution shift \citep{shi2023how}, while also capturing better semantic information \citep{assran2023self} compared to representations learned through supervised objectives. Moreover, self-supervised graph representations have demonstrated superior performance on downstream tasks compared to representations learned in a fully supervised manner, indicating better generalization capabilities \citep{Hu2020Strategies, Sun2020InfoGraph, you2020graph, DBLP:conf/cikm/FuXLTH20, you2021graph, pmlr-v119-hassani20a, velickovic_2019_iclr, zhu2020deep}. Additionally, multi-task learning with self-supervised objectives has been shown to achieve better performance on downstream tasks \citep{doersch2017multi, ghiasi2021multi}. To leverage these benefits, we propose training the GNN encoder with three self-supervised objectives. Unlike RQ-VAE \citep{lee2022autoregressive}, which uses reconstruction as its primary objective, we employ graph-specific objectives to capture the nuances of both structure and features.  

\begin{figure}[t]
    \centering
    \includegraphics[width=5.9in]{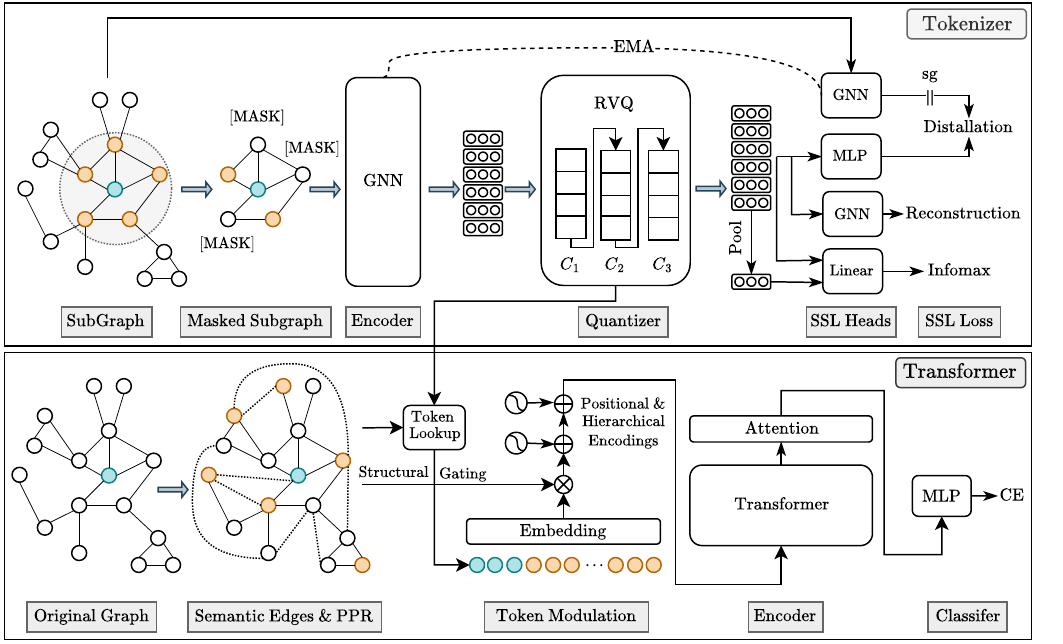}
    \caption{\small Overview of our proposed Graph Quantization Transformer (GQT) consisting of three main components: (1) a GNN to encode local interactions, (2) vector quantization for compact representation, and (3) generative and contrastive heads for robust representation learning. We also utilize a Transformers encoder to model long-range interactions. We augment the graph with semantic edges (dashed lines) and generate a sequence for each node based on Personalized PageRank scores. We then modulate the tokens through hierarchical encoding and structural gating, and feed them into the Transformer and aggregate the learned representations through an attention module before passing it to the classification head.}
    \label{fig:overview}
\end{figure}

\subsection{Training}
To capture different aspects of information, we employ a multi-task learning framework that leverages three distinct families of graph self-supervised objectives: student-teacher distillation \citep{thakoor2022largescale}, masked autoencoding \citep{hou2022graphmae}, and Infomax \citep{velickovic_2019_iclr}. We also introduce a commitment loss \citep{van2017neural} to enforce alignment between learned node representations and the codebook representations.
Specifically, the GNN encoder is trained through gradient descent to minimize a loss function comprising of three terms, where $\beta$ is the loss weight:
\begin{equation}
    \mathcal{L} = \mathcal{L}_\text{dgi} + \mathcal{L}_\text{gmae2} + \beta \mathcal{L}_\text{commit}
\end{equation}
The first term is the Deep Graph Infomax (DGI) \citep{velickovic_2019_iclr} objective, which maximizes mutual information (MI) between node representations and graph (sub-graph) representations, based on the Jensen-Shannon divergence between the joint and product of marginals as follows:
\begin{equation}
    \mathcal{L}_\text{dgi} = \mathbb{E} \left(\sum\limits_{v\in g} \log\left(\mathcal{D}\left(h_v, h_g\right)\right) + 
    \sum\limits_{u\in \tilde{g}} \log\left(1-\mathcal{D}\left(h_u, h_g\right)\right) \right)
\end{equation}
where $h_u$ is the representation of node $u$. $h_g$ is the global (sub-graph/graph) representation, computed as the mean of node representations. $\tilde{g}$ is the corrupted version of the original graph, with the same structure but randomly shuffled features, providing negative examples for contrastive learning. $\mathcal{D}\left(h_u, h_g\right)=\sigma\left(h_u^T\textbf{W}h_g \right)$ is the discriminator that scores whether a node belongs to the graph, and is defined as a bilinear classifier.

The second term is the GraphMAE2 objective \citep{hou2023graphmae2}, which combines the generative loss of GraphMAE \citep{hou2022graphmae} with the teacher-(noisy)student distillation loss of BGRL \citep{thakoor2022largescale}. This combination enables the model to avoid overfitting and learn more semantic representations. The GraphMAE2 loss is computed as follows:
\begin{equation}
    \mathcal{L}_\text{gmae2} = \sum\limits_{v \in \tilde{g}}\left(1-\frac{x_v^T.\tilde{h}_v}{\Vert x_v^T\Vert.\Vert\tilde{h}_v\Vert} \right)^\gamma + 
    \lambda \sum\limits_{v \in g}\left( 1-\frac{h_v^T.\tilde{h}_v}{\Vert h_v^T\Vert.\Vert\tilde{h}_v\Vert}\right)^\gamma
\end{equation}
where $\tilde{g}$ is the masked graph, $\tilde{h}_v$ is the node representation of a masked node learned by the noisy student, $h_v$ is the corresponding node representation learned by the teacher over the original graph, and $\gamma \geq 1$ is a scaling factor. The teacher's parameters are updated using an Exponential Moving Average (EMA) of the noisy student's parameters. 

The third term is the commitment loss, which encourages the representations to get close to their corresponding codebook embeddings within the RVQ layer. This loss is computed as:
\begin{equation}
    \mathcal{L}_\text{commit} = \frac{1}{|\mathcal{V}|}\sum\limits_{v\in g} ||h_v - \text{sg}\left[c_k \right] ||_2
\end{equation}
where sg is the stop-gradient operator, and $c_k$ is the representation of the codebook that $h_v$ is assigned to (i.e., the centroid or prototype vector). Note that this loss only affects the  node representations and does not update the codebooks.

To initialize and update the codebooks, we employ K-Means clustering and EMA with weight decay $\tau \in [0, 1]$, respectively. Specifically, the codebooks are updated as follows:
\begin{equation}
    c_k^{t} = \tau c_k^{t-1} + (1-\tau) \frac{1}{|\mathcal{V}_k|}\sum\limits_{v \in \mathcal{V}_k} h_v
\end{equation}
where $\mathcal{V}_k$ is the set of nodes assigned to codebook $c_k$. This update rule allows the codebooks to adapt to the changing node representations while maintaining stability.

\section{Graph Transformer} \label{sec:transformer}
\subsection{Graph Serialization}
Once the tokenizer is trained, each node $v \in \mathcal{V}$ is mapped to $c$ discrete tokens: $T_{v}=\left[t^v_1, \cdots,t^v_c\right] \in \mathbb{N}^{c}$ (i.e., $T_{v}=\textbf{X}_Q[v]$), encoding local interactions of that node. We then need to serialize the graph in order to input it to the Transformer.

\textbf{Semantic Edges}. To enable the Transformer to capture long-range interactions, the input should consist of a sequence of tokens from nodes that are likely to have long-range dependencies. To facilitate this, we first augment the graph with \emph{semantic edges} denoted as $\mathcal{E}_s$, which are computed as follows:
\begin{equation}
    \mathcal{E}_s = \left\{e_{u,v} \mid \argtopk\limits_{u \in \mathcal{V}} \text{sim} \left(f\left(x_u\right), f\left(x_v\right)\right) \forall v \in \mathcal{V}\right\}
\end{equation}
where $\text{sim}(\cdot, \cdot)$ denotes the similarity function, $x_u$ is the feature vector of node $u$, and $f$ is a projection function. We use cosine similarity as the similarity function  and Principal Component Analysis (PCA) as the projection function. The semantic edge augmentation effectively creates sparse edges between each node and its k-nearest neighbors in the feature space, enhancing the model's ability to recognize and utilize long-range dependencies. 

\textbf{Structural Serialization}. We combine the semantic edges with the original edges and use Personalized PageRank (PPR) to generate a sequence per target node. This enriches the sequence with information beyond local interactions, allowing the Transformer to access potential long-range dependencies. We construct the sequence $S_v$ for each node $v$ as follows: 
\begin{equation}
    S_v = \left[T_v \Vert T_u\Vert_{u \in \argtopk \text{PPR}\left(v, \mathcal{E} \cup \mathcal{E}_s\right)}\right]
\end{equation}

where $S_v=\left[t^v_1 \cdots t^v_c \mid t^{u_1}_1\cdots t^{u_1}_c\mid \cdots \mid  t^{u_k}_1 \cdots t^{u_k}_c\right]$ is a sequence of length $c\times (k+1)$, comprising discrete tokens that represent the target node $v$, followed by discrete tokens of the top-$k$ relevant nodes to node $v$. These relevant nodes are determined based on PPR scores. Note that the computation of semantic edges and PPR sequences is performed only once as a pre-processing step, thereby reducing the computational overhead during training.

\subsection{Token Modulation}
\textbf{Token Embeddings}. There are $c \times K$ possible discrete tokens, where $c$ is the number of codebooks and $K$ is the codebook size. We randomly initialize an embedding matrix $\textbf{X}_T \in \mathbb{R}^ {c \times K \times d_x}$, which is trained end-to-end alongside the Transformer. To further enrich the token representations, we introduce an additional token for each node that aggregating the embeddings of its assigned codebooks from the pretrained tokenizer:
\begin{equation}
    h_c^v = \sum\limits_{i=1}^{c}\textbf{C}[i, t_i^v]
\end{equation}
where $\textbf{C}[i, j]$ is the embedding corresponding to index j in the ith codebook. We found that adding this explicit aggregated token from the codebook leads to better performance compared to initializing $\textbf{X}_T$ directly with $\textbf{C}$. The input representation of the sequence for node $v$ is then defined as: 
\begin{equation}
S_v=\left[\textbf{X}_T[i, t^v_i]\operatorname*{\Vert}_{i=1}^c h_c^v \ \Vert \ \textbf{X}_T[i, t^{u_1}_i]\operatorname*{\Vert}_{i=1}^c \  h_c^{u_1}\ \Vert \cdots\Vert \textbf{X}_T[i, t^{u_k}_i]\operatorname*{\Vert}_{i=1}^c \  h_c^{u_k}\right]
\end{equation}
where $[\textbf{X}||\textbf{Y}]$ denotes concatenation of sequences $\textbf{X}$ and $\textbf{Y}$. This representation combines the individual token embeddings with the aggregated codebook embeddings, providing a more comprehensive and nuanced input to the Transformer.

\textbf{Structural Gating}. In order to provide the Transformer with the global structural importance scores of the nodes within the sequence with respect to the target node, we introduce a gating mechanism over the input token embeddings as follows:
\begin{equation}
    S_v = S_v \odot \text{Softmax}\left(\topk \text{PPR}\left(v, \mathcal{E} \cup \mathcal{E}_s\right)\right)
\end{equation}
where we first apply a softmax  function with temperature $\tau=1$ to normalize the PPR scores, and then multiply each node token's representation by its corresponding normalized score. 

\textbf{Positional Encoding}. We also introduce two trainable positional encodings to the input tokens. The first positional encoding enables the Transformer to distinguish between tokens from different nodes,  while the second encoding, referred to as hierarchical encoding, allows the Transformer to recognize the hierarchy level of each token within the codebooks. We randomly initialize the  positional encodings $\textbf{PE} \in \mathbb{R}^{(k+1)\times d_x}$ and $\textbf{HE} \in \mathbb{R}^{c\times d_x}$ and sum them with the encoding of their corresponding token. For example, the final encoding of the token $j$  of the node $i$ within the sequence is computed as: $x=\textbf{X}_T[j, t^{u_i}_j]+\textbf{PE}[i] + \textbf{HE}[j]$. Note that we did not use any structural encoding, such as Laplacian eigenvectors, as we did not observe any significant gains from them.

\subsection{Transformer Encoder}
We use $l$ layers of standard Transformer encoder with flash attention \citep{dao2022flashattention} to generate contextual representations per token in the sequence: $\textbf{H}^{(l)}\in \mathbb{R}^{(c+1)\times(k+1)\times d_h}$. We then aggregate the token representations for $j$-th node in the sequence by summing along the token dimension:
\begin{equation}
    \textbf{H}_{v_j}=\sum\limits_{i=1}^{c+1}\textbf{H}^{(l)}[i, j]\in \mathbb{R}^{(k+1)\times d_h}
\end{equation}
To obtain a single representation for the entire sequence, We further aggregate the representation using a linear attention layer:
\begin{equation}
    h=\sum\limits_{i=1}^{k+1}\alpha_ih_i \quad \text{where} \quad \alpha_i=\frac{\exp(\textbf{W}h_i)}{\sum_j\exp(\textbf{W}h_j)}
\end{equation}
We feed the resulting representation into a fully-connected classifier and train the model end-to-end using cross-entropy loss. Note that during inference, only the Transformer and classifier are utilized, as the tokenizer is pretrained and the sequences are pre-computed. Furthermore, since we only require discrete tokens and codebook embeddings, our approach enables efficient memory usage,  regardless of graph size, allowing for efficient training and inference on large-scale graphs.

\section{Experiments}
We evaluate GQT on both medium- and large-scale graph learning tasks, encompassing 22 homophilic, heterophilic, and long-range benchmarks. We follow the established experimental protocols from previous works to ensure fair comparisons. Details of the datasets, experimental setup, and hyperparameters are provided in Appendices \ref{appendix:datasets} and \ref{appendix:hyperparameters}, respectively.

\subsection{Comparison with State-of-the-Art}\label{sec:main_results}

\textbf{Long-Range Benchmarks}. We use four datasets from the Long-Range Graph Benchmark (LRGB)~\citep{dwivedi2022long}, including the Peptides-Func dataset for graph classification with Average Precision (AP) metric, the Peptides-Struct dataset for graph regression with Mean Absolute Error (MAE) metric, the COCO-SP dataset for inductive node classification with macro F1 metric, and the PCQM-Contact for link prediction with Mean Reciprocal Rank (MRR) metric. We compare our results to baselines reported in \citep{wang2024graph}. The results shown in Table \ref{tab:LRGB} suggest that GQT is able to capture long-range dependencies and performs well on various graph prediction tasks.

\begin{table}[ht]
    \centering
    \caption{Mean performance on inductive long-range benchmarks over five runs.}
    \resizebox{0.75\textwidth}{!}{
    \setlength{\tabcolsep}{3pt}
    \begin{tabular}{lcccc}\toprule
    \textbf{Task} &\textbf{Graph Classification} &\textbf{Graph Regression} &\textbf{Node Classification} &\textbf{Link Prediction} \\\midrule
    Dataset &Peptides-Func &Peptides-Struct &COCO-SP &PCQM-Contact \\
    \#Graphs & 15,535 & 15,535 & 123,286 & 529,434 \\
    Avg. \#Nodes & 150.94 & 150.94 & 476.88 & 30.14 \\
    Avg. \#Edges & 307.30 & 307.30 & 2,693.67 & 61.09 \\
    Metric &AP ↑ &MAE ↓ &F1 ↑ &MRR ↑ \\\midrule
    GCN &0.5930$\pm$0.0023 &0.3496$\pm$0.0013 &0.0841$\pm$0.0010 &0.3234$\pm$0.0006 \\
    Exphormer &0.6258$\pm$0.0092 &0.2512$\pm$0.0025 &0.3430$\pm$0.0108 &\textbf{0.3587$\pm$0.0025} \\
    GPS &0.6535$\pm$0.0041 &0.2500$\pm$0.0005 &0.3412$\pm$0.0044 &0.3337$\pm$0.0006 \\
    Graph-Mamba &0.6739$\pm$0.0087 &0.2478$\pm$0.0016 &0.3960$\pm$0.0175 &0.3395$\pm$0.0013 \\
    GQT (Ours) &\textbf{0.6903$\pm$0.0085} &\textbf{0.2452$\pm$0.0018} &\textbf{0.4007$\pm$0.0125} &0.3427$\pm$0.0012 \\
    \bottomrule
    \end{tabular}}
    \label{tab:LRGB}
\end{table}

\begin{table}[t]\centering
\caption{\small Mean node classification accuracy on medium-scale homophilic datasets over five runs.}
\setlength{\tabcolsep}{3pt}
\label{tab:results_small_homophilic}
\resizebox{\textwidth}{!}{
\begin{tabular}{llcccccccc}
\toprule
\multirow{7}{*}{{\begin{turn}{90}Dataset\end{turn}}}
& & \textbf{CoraFull} &\textbf{CiteSeer}&\textbf{PubMed} &\textbf{Computer} &\textbf{Photo} &\textbf{CS} &\textbf{Physics}  &\textbf{WikiCS}\\
\cmidrule{2-10}
& \#Nodes & 19,793 & 3,327 & 19,717 & 13,752 & 7,650 & 18,333 & 34,493 & 11,701 \\
& \#Edges & 126,842 & 4,522 & 88,651 & 491,722 & 238,163 & 163,788 & 495,924 & 216,123 \\
& \#Features & 8,710 & 3,703 & 500 & 767 & 745 & 6,805 & 8,415 & 300 \\
& \#Classes & 70 & 6 & 3 & 10 & 8 & 15 & 5 & 10 \\
\midrule
\multirow{8}{*}{{\begin{turn}{90}\textbf{GNN}\end{turn}}} 
&GCN & 61.76$\pm$0.14 & 76.50$\pm$1.36 & 86.54$\pm$0.12 &89.65$\pm$0.52 &92.70$\pm$0.20 &92.92$\pm$0.12 &96.18$\pm$0.07 & 77.47$\pm$0.85\\
&GAT & 64.47$\pm$0.18 &76.55$\pm$1.23&86.32$\pm$0.16 &90.78$\pm$0.13 &93.87$\pm$0.11 &93.61$\pm$0.14 &96.17$\pm$0.08  & 76.91$\pm$0.82\\
&APPNP & 65.16$\pm$0.28 &76.53$\pm$1.16&88.43$\pm$0.15 &90.18$\pm$0.17 &94.32$\pm$0.14 &94.49$\pm$0.07 &96.54$\pm$0.07  & 78.87$\pm$0.11\\
&GPRGNN & 67.12$\pm$0.31 &77.13$\pm$1.67&89.34$\pm$0.25 &89.32$\pm$0.29 &94.49$\pm$0.14 &95.13$\pm$0.09 &96.85$\pm$0.08  &78.12$\pm$0.23\\
&GraphSAINT & 67.85$\pm$0.21 &$-$&88.96$\pm$0.16 &90.22$\pm$0.15 &91.72$\pm$0.13 &94.41$\pm$0.09 &96.43$\pm$0.05  &$-$\\
&GraphSAGE & $-$ & 75.58$\pm$1.33 & 87.48$\pm$0.38 &91.20$\pm$0.29 &94.59$\pm$0.14 &93.91$\pm$0.13 &96.49$\pm$0.06 & 74.77$\pm$0.95\\
&PPRGo & 63.54$\pm$0.25 &$-$&87.38$\pm$0.11 &88.69$\pm$0.21 &93.61$\pm$0.12 &92.52$\pm$0.15 &95.51$\pm$0.08  &78.12$\pm$0.23\\
&GRAND+ & 71.37$\pm$0.11 &$-$&88.64$\pm$0.09 &88.74$\pm$0.11 &94.75$\pm$0.12 &93.92$\pm$0.08 &96.47$\pm$0.04  &$-$\\
\midrule
\multirow{11}{*}{{\begin{turn}{90}\shortstack[c]{\textbf{GT}}\end{turn}}}
& GT & 61.05$\pm$0.38 &$-$&88.79$\pm$0.12 &91.18$\pm$0.17 &94.74$\pm$0.13 &94.64$\pm$0.13 &97.05$\pm$0.05  &$-$\\
&Graphormer & OOM &$-$&OOM &OOM &92.74$\pm$0.14 &94.64$\pm$0.13 &OOM  &$-$\\
&SAN & 59.01$\pm$0.34 &$-$&88.22$\pm$0.15 &89.93$\pm$0.16 &94.86$\pm$0.10 &94.51$\pm$0.15 &OOM  &$-$\\
&GraphGPS & 55.76$\pm$0.23 &76.99$\pm$1.12&88.94$\pm$0.16 &OOM &95.06$\pm$0.13 &93.93$\pm$0.15 &OOM  &78.66$\pm$0.49\\
 & GOAT       & $-$ & 76.89$\pm$1.19 & 86.87$\pm$0.24 & 90.96$\pm$0.90 & 92.96$\pm$1.48 & 94.21$\pm$0.38 & 96.24$\pm$0.24 & 77.00$\pm$0.77 \\
 & NodeFormer &$-$ & 76.33$\pm$0.59 & 89.32$\pm$0.25  & 86.98$\pm$0.62 & 93.46$\pm$0.35 & 95.64$\pm$0.22 & 96.45$\pm$0.28 &74.73$\pm$0.94 \\
 & DIFFormer  & $-$ & 76.72$\pm$0.68 & 89.51$\pm$0.67  & 91.99$\pm$0.76& 95.10$\pm$0.47  & 94.78$\pm$0.20 & 96.60$\pm$0.18 & 73.46$\pm$0.56 \\
 & NAGphormer & 71.51$\pm$0.13 &77.42$\pm$1.41&89.70$\pm$0.19 &91.22$\pm$0.14 &95.49$\pm$0.11 &95.75$\pm$0.09 &97.34$\pm$0.03  &77.16$\pm$0.72\\
 & Exphormer & 69.09$\pm$0.72 &76.83$\pm$1.24&89.52$\pm$0.54 &91.59$\pm$0.31 &95.27$\pm$0.42 &95.77$\pm$0.15 &97.16$\pm$0.13  &78.54$\pm$0.49\\
 & VCR-Graphormer & 71.67$\pm$0.10 &$-$&89.77$\pm$0.15 &91.75$\pm$0.15 &95.53$\pm$0.14 &95.37$\pm$0.04 &97.34$\pm$0.04  &$-$\\
\cmidrule{2-10}
&{GQT (ours)} & \textbf{71.81$\pm$0.21} &\textbf{77.84$\pm$0.94} &\textbf{90.14$\pm$0.16} &\textbf{93.37$\pm$0.44} &\textbf{95.73$\pm$0.18} &\textbf{96.11$\pm$0.09} &\textbf{97.53$\pm$0.06}  &\textbf{80.14$\pm$0.57}\\
\bottomrule
\end{tabular}}
\end{table}

\textbf{Homophilic Node Classification.}
We use eight medium-scale homophilic datasets including: CoraFull \citep{bojchevski2017deep}, CiteSeer, PubMed \citep{yang2016revisiting}, Amazon Computers, Amazon Photos, Co-author CS, Co-author Physics \citep{shchur2018pitfalls}, and WikiCS \citep{mernyei2020wiki}. We compare our results with eight GNNs including: GCN \citep{kipf2017semisupervised}, GAT, APPNP \citep{gasteiger2018predict}, GPRGNN \citep{chien2020adaptive}, GraphSAINT \citep{zeng2019graphsaint}, GraphSAGE \citep{hamilton2017inductive}, PPRGo \citep{bojchevski2020scaling}, and GTAND+ \citep{feng2022grand+}. We also compare against ten GTs including GT \citep{dwivedi2020generalization}, Graphormer \citep{ying2021transformers}, SAN \citep{kreuzer2021rethinking}, GraphGPS \citep{rampavsek2022recipe}, GOAT \citep{kong2023goat}, NodeFormer  \citep{wu2022nodeformer}, DiffFormer \citep{wu2023difformer}, NAGphormer \citep{chen2023nagphormer}, Exphormer \citep{pmlr-v202-shirzad23a}, and VCR-Graphormer \citep{fu2024vcrgraphormer}. The baseline performance is reported from \citep{wu2023simplifying, luo2024structure}. GQT outperforms the baseline GNN and GT models on 7 out of 8 benchmarks (Table \ref{tab:results_small_homophilic}). Notably, this achievement comes with a significant memory reduction. For example, on the Physics dataset with 34,493 nodes, we only use $256 \times 6$ tokens, i.e., a 23-fold memory reduction.  

\textbf{Heterophilic Node Classification.} 
We also evaluate GQT on six medium-scale heterophilic datasets: Squirrel, Chameleon \citep{rozemberczki2021multi}, Questions, Roman-Empire, Amazon-Ratings, and Minesweeper \citep{platonov2023critical}. We compare the performance with seven GNNs: GCN, GraphSAGE, GAT, GPRGNN, H2GCN \citep{zhu2020beyond}, CPGNN \citep{zhu2021graph}, and GloGNN \citep{li2022finding}, and six GTs: GraphGPS, GOAT, NodeFormer, SGFormer, NAGphormer, and Exphormer. The baseline performance is reported from \citep{wu2023simplifying, luo2024classic, platonov2023critical, behrouz2024graph}. As shown in Table~\ref{tab:results_heterophilic}, GQT outperforms the baselines on five out of six datasets. We observe that introducing semantic edges and structural gating specifically benefits the heterophilic setting (Appendix \ref{appendix:ablation}), as they enable the Transformer to capture long-range dependencies that are not easily accessible through the original graph structure.

\textbf{Large-scale Node Classification}
We also use four large-scale datasets: ogbn-proteins, ogbn-arxiv, ogbn-products \citep{hu2020open}, and pokec (heterogeneous) \citep{leskovec2016snap}. We compare the performance against six GNN:  LINKX \citep{lim2021large}, SIGN \citep{frasca2020sign}, GCN, GAT, GraphSAGE, and GPRGNN; and six GTs: GraphGPS, GOAT, NodeFormer, NAGphormer, Exphormer, and SGFormer \citep{wu2023simplifying}. We report the baseline performance from \citep{wu2023simplifying, luo2024structure}. The results  (Table~\ref{tab:results_large}) show that GQT outperforms the  baseline models on all large-scale benchmarks. This achievement comes with a significant reduction in required memory. For instance, on the ogbn-products dataset with 2,449,029 nodes and 100-dimensional node features, GQT requires only 3 codebooks of size 4096, resulting in a 30-fold memory reduction.

\begin{table}[t]
\small
\setlength{\tabcolsep}{4pt}
\caption{\small Mean node classification performance on heterophilic graphs over five runs.}
\label{tab:results_heterophilic}
\begin{center}
\resizebox{0.9\textwidth}{!}{
\begin{tabular}{llcccccc}
\toprule
\multirow{7}{*}{{\begin{turn}{90}Dataset\end{turn}}}
 & & \textbf{Squirrel} & \textbf{Chameleon} &\textbf{Amazon-Ratings} &\textbf{Roman-Empire} &\textbf{Minesweeper} &\textbf{Questions} \\
 \cmidrule{2-8}
 & \#Nodes & 5,201  &2,277  &22,662 & 24,492 & 10,000 & 48,921 \\
 & \#Edges & 216,933  &36,101  &32,927 & 93,050 & 39,402 & 153,540 \\
 & \#Features & 2,089 & 2,325 & 300 & 300 & 7 & 301\\
 & \#Classes & 5 & 5 & 18 & 5 & 2 & 2 \\
 & Measure & Accuracy$\uparrow$ & Accuracy$\uparrow$ & Accuracy$\uparrow$ &Accuracy$\uparrow$ &ROC-AUC$\uparrow$&ROC-AUC$\uparrow$  \\
 \midrule 
\multirow{7}{*}{{\begin{turn}{90}GNN\end{turn}}} 
& GCN & 38.67$\pm$1.84& 41.31$\pm$3.05&48.70$\pm$0.63&73.69$\pm$0.74&  89.75$\pm$0.52& 76.09$\pm$1.27 \\
& GraphSAGE & 36.09$\pm$1.99& 37.77$\pm$4.14&53.63$\pm$0.39&85.74$\pm$0.67 &93.51$\pm$0.57 &76.44$\pm$0.62\\
& GAT & 35.62$\pm$2.06&39.21$\pm$3.08  &52.70$\pm$0.62&88.75$\pm$0.41 &93.91$\pm$0.35 & 76.79$\pm$0.71\\
& H2GCN &  35.10$\pm$1.15& 26.75$\pm$3.64& 36.47$\pm$0.23&60.11$\pm$0.52 &89.71$\pm$0.31 &63.59$\pm$1.46 \\ 
& CPGNN &  30.04$\pm$2.03& 33.00$\pm$3.15& 39.79$\pm$0.77&63.96$\pm$0.62 &52.03$\pm$5.46 &65.96$\pm$1.95 \\ 
& GPRGNN&  38.95$\pm$1.99& 39.93$\pm$3.30& 44.88$\pm$0.34&64.85$\pm$0.27 &86.24$\pm$0.61 &55.48$\pm$0.91 \\ 
& GloGNN&  35.11$\pm$1.24& 25.90$\pm$3.58& 36.89$\pm$0.14&59.63$\pm$0.69 &51.08$\pm$1.23 &65.74$\pm$1.19  \\ 
 \midrule 
\multirow{6}{*}{{\begin{turn}{90}GT\end{turn}}} 
& GraphGPS& 39.67$\pm$2.84&40.79$\pm$4.03   & 53.10$\pm$0.42&82.00$\pm$0.61 &90.63$\pm$0.67 &71.73$\pm$1.47 \\ 
& NodeFormer&  38.52$\pm$1.57& 34.73$\pm$4.14&  43.86$\pm$0.35&64.49$\pm$0.73& 86.71$\pm$0.88& 74.27$\pm$1.46\\ 
& SGFormer&  41.80$\pm$2.27& \textbf{44.93$\pm$3.91}&  48.01$\pm$0.49&79.10$\pm$0.32& 90.89$\pm$0.58& 72.15$\pm$1.31 \\
& NAGphormer &35.80$\pm$1.33 &$-$ & 51.26$\pm$0.72 &74.34$\pm$0.77 &84.19$\pm$0.66 &$-$ \\
& Exphormer &36.04$\pm$1.45 &$-$ & 53.51$\pm$0.46 & 89.03$\pm$0.37 &90.74$\pm$0.53 &$-$ \\ 
\cmidrule{2-8}
& GQT(ours) &\textbf{42.72$\pm$1.69} &44.23$\pm$3.05  &\textbf{54.32$\pm$0.41}& \textbf{90.98$\pm$0.24} &\textbf{97.36$\pm$0.35} & \textbf{78.94$\pm$0.86} \\
\bottomrule
\end{tabular}}
\end{center}
\end{table}

\begin{table}[t]\centering
\caption{\small Mean node classification performance on large-scale datasets over five runs.}
\label{tab:results_large}
\setlength{\tabcolsep}{4.5pt}
\begin{center}
\resizebox{0.65\textwidth}{!}{
\begin{tabular}{llcccc}
\toprule
\multirow{7}{*}{{\begin{turn}{90}Dataset\end{turn}}}
 & & \textbf{ogbn-proteins}&\textbf{ogbn-arxiv} &\textbf{ogbn-products} &\textbf{pokec}\\
 \cmidrule{2-6}
 & \#Nodes & 132,534  &169,343  &2,449,029 & 1,632,803 \\
 & \#Edges & 39,561,252  &1,166,243  &61,859,140 & 30,622,564 \\
 & \#Features & 128 & 8 & 100 & 65 \\
 & \#Classes & 40 & 2 & 47 & 2 \\
 & Measure &  ROC-AUC$\uparrow$&Accuracy $\uparrow$ & Accuracy $\uparrow$ & Accuracy $\uparrow$ \\ 
\midrule
\multirow{7}{*}{{\begin{turn}{90}\textbf{GNN}\end{turn}}}
 & GCN & 72.51$\pm$0.35 & 71.74$\pm$0.29 & 75.64$\pm$0.21  &75.45$\pm$0.17 \\
 & GAT & 72.02$\pm$0.44 & 71.95$\pm$0.36 & 79.45$\pm$0.59 & 72.23$\pm$0.18 \\
 & GPRGNN & 75.68$\pm$0.49 & 71.10$\pm$0.12 & 79.76$\pm$0.59 & 72.23$\pm$0.18  \\
 & LINKX & 71.37$\pm$0.58 & 66.18$\pm$0.33 & 71.59$\pm$0.71 & 82.04$\pm$0.07\\
 & GraphSAGE & 77.68$\pm$0.20&71.49$\pm$0.27 &78.29$\pm$0.16 & 75.63$\pm$0.38\\
 & SIGN &$-$ &71.95$\pm$0.11 &80.52$\pm$0.16 &$-$ \\
\midrule
\multirow{8}{*}{{\begin{turn}{90}\textbf{GT}\end{turn}}}
  & GraphGPS & 76.83$\pm$0.26 & 70.97$\pm$0.41 & OOM & OOM  \\
  & GOAT   & 74.18$\pm$0.37 & 72.41$\pm$0.40 & 82.00$\pm$0.43 & 66.37$\pm$0.94 \\
  & NodeFormer & 77.45$\pm$1.15 & 59.90$\pm$0.42 & 72.93$\pm$0.13 &71.00$\pm$1.30 \\
  & SGFormer   & 79.53$\pm$0.38 & 72.63$\pm$0.13 & 74.16$\pm$0.31 &73.76$\pm$0.24 \\
  & NAGphormer & 73.61$\pm$0.33  & 70.13$\pm$0.55 & 73.55$\pm$0.21 &76.59$\pm$0.25\\
  & Exphormer  & 74.58$\pm$0.26  & 72.44$\pm$0.28 & OOM & OOM   \\
 \cmidrule{2-6}
  & GQT(ours) &\textbf{82.13$\pm$0.34} &\textbf{73.14$\pm$0.16} &\textbf{82.46$\pm$0.17} &\textbf{83.76$\pm$0.24} \\
\bottomrule
\end{tabular}}
\end{center}
\end{table}

\subsection{Ablation Study}
\textbf{Effect of Tokenization}. We examine the performance of the tokenizer by training a linear model on the representations of the learned tokens without modulation, augmentation, or Transformer (1). As shown in Table~\ref{tab:ablation}, within the linear evaluation protocol, the tokenizer shows strong performance, surpassing that of GTs such as GraphGPS and NAGphormer, as well as GNNs like GAT and SIGN (Table \ref{tab:results_large}). This implies that the tokenizer is capable of learning effective token representations. To further investigate the importance of the tokenizer, we exclude it and train the Transformer directly on the original node features (2). As expected, this results in significant degradation in performance, highlighting the crucial role of the tokenizer. Additionally, to study the effects of vector quantization, GraphMAE2, and DGI objectives, we train the model by excluding each component (3-5). The results suggest that the SSL objectives contribute more significantly to the performance compared to vector quantization. This is because the primary purpose of vector quantization is to compress information into discrete tokens, reducing memory requirements. Between GraphMAE2 and DGI, the former yields the highest gain. This is due to its composition of two objectives: masked reconstruction and teacher-(noisy)student distillation. Both of these objectives have been shown to outperform InfoMax objectives on downstream tasks \citep{hou2022graphmae, thakoor2022largescale}.

\textbf{Effect of Modulation}. We also investigate the impact of codebook embeddings, positional encoding, and structural gating on the model's performance (6-8). As shown in Table~\ref{tab:ablation}, introducing aggregated codebook embeddings leads to improved downstream performance because it provides the Transformer with richer representations of each token. Positional encoding, as observed in other domains, contributes moderately to downstream performance. We also note that introducing structural gating yields moderate improvements in homophilic settings, whereas the gains are significant in heterophilic benchmarks (\ref{appendix:ablation}). This disparity can be attributed to the ability of structural gating to provide the Transformer with importance scores computed over the global graph structure, which is particularly beneficial in heterophilic scenarios.

\textbf{Effect of Augmentation}. We study the effect of semantic edges on downstream performance (9). The results suggest that augmenting the graph structure with semantic edges yields significant gains. This is because introducing semantic edges allows the Transformer to access semantic information that may not be captured by the original graph structure. Furthermore, when combined with random walks, this also enables the Transformer to attend to long-range dependencies which is particularly important in heterophilic benchmarks, where semantic relationships between nodes are more nuanced.


\begin{table}[t]
    \setlength{\tabcolsep}{3pt}
    \centering
    \caption{Accuracy drop under adversarial attack.}
    \resizebox{0.5\textwidth}{!}{
    \begin{tabular}{lcccc}\toprule
    Attack & \multicolumn{2}{c}{GR-BCD} & \multicolumn{2}{c}{PR-BCD}\\\midrule
    & PubMed & ogbn-arxiv & PubMed & ogbn-arxiv \\\midrule
    RQ-VAE &   20.40\% & 14.80\% &23.30\% &17.20\% \\
    GQT (ours) & 15.80\% &10.40\% &18.10\% &11.30\% \\
    \bottomrule
    \end{tabular}}
    \label{tab:robust}
\end{table}
\textbf{Robustness Analysis}. To measure robustness, we use Greedy Randomized Block Coordinate Descent (GRBCD) and Projected Randomized Block Coordinate Descent (PR-BCD) \citep{geisler2021robustness} adversarial attacks to measure the accuracy degradation. We compare the GQT with an RQ-VAE~\citep{lee2022autoregressive}. The results (Table \ref{tab:robust}) show that our tokenizer is more robust to attacks. This is because GQT is trained with multi-task self-supervised objectives while RQ-VAE is trained with reconstruction objective. More details are provided in Appendix \ref{appendix:additional_results}.

\begin{table}[t]
\caption{\small Ablation study on effect of proposed components on the ogbn-arxiv dataset.}
\label{tab:ablation}
\begin{center}
\setlength{\tabcolsep}{3pt}
\resizebox{\textwidth}{!}{
\begin{tabular}{lccc ccc cc c c}\toprule
 & \multicolumn{3}{c}{\textbf{Graph Tokenizer}} & \multicolumn{3}{c}{\textbf{Token Modulation}} & \multicolumn{2}{c}{\textbf{Augmentation}} &\textbf{Model} & \textbf{Performance}\\
 \cmidrule{2-9}\cmidrule{11-11}
 & RVQ	& GMAE2	& DGI & Codebook   & Positional & Structural & Semantic & PPR      & & Accuracy$\uparrow$ \\
 &     &	    &	  &	Embeddings & Encoding	& Gating	 & Edges	& Sequence & & \\
\midrule
(1) & \cmark &\cmark &\cmark & \cmark & & & & & Linear & 71.91$\pm$0.13 \\
(2) &  &  & & &\cmark & & &\cmark & Transformer & 70.68$\pm$0.17 \\
\midrule
(3) & &\cmark &\cmark &\cmark & \cmark & \cmark&\cmark & \cmark & Transformer & 72.84$\pm$0.23 \\
(4) & \cmark & &\cmark &\cmark & \cmark & \cmark&\cmark & \cmark & Transformer & 71.83$\pm$0.19 \\
(5) & \cmark & \cmark & &\cmark & \cmark & \cmark &\cmark & \cmark & Transformer & 72.71$\pm$0.24 \\
\midrule
(6) & \cmark &\cmark &\cmark &  &\cmark & \cmark& \cmark &\cmark & Transformer & 71.34$\pm$0.16 \\
(7) & \cmark  &\cmark &\cmark &\cmark &  & \cmark &\cmark & \cmark & Transformer &  72.69$\pm$0.21 \\
(8) & \cmark & \cmark & \cmark &\cmark & \cmark & &\cmark & \cmark & Transformer &  73.08$\pm$0.14 \\
\midrule
(9) & \cmark &\cmark &\cmark &\cmark &\cmark &\cmark & &\cmark & Transformer & 72.59$\pm$0.25 \\
\midrule
(10) & \cmark & \cmark & \cmark &\cmark & \cmark & \cmark &\cmark & \cmark & Transformer & \textbf{73.14$\pm$0.16} \\
\bottomrule
\end{tabular}}
\end{center}
\end{table}

\newpage
\section{Conclusion}
We introduced the \textbf{G}raph \textbf{Q}uantized \textbf{T}okenizer (GQT) to provide standard Transformer encoders to acces discrete graph tokens that encapsulate local interactions and allow Transformers to attend to long-range dependencies within the graph structure. This allows us to seamlessly take advantage of the rapid advances in scaling Transformers. We achieved state-of-the-art performance on 20 out of 22 datasets, including large-scale and long-range homophilic and heterophilic datasets. As future directions, we plan to explore the potential of GQT in generative graph learning. Additionally, we aim to couple GQT with LLMs to provide a shared feature space across various graph datasets, paving the way for true Graph Foundational Models (GFMs) \citep{liu2023towards, mao2024position}. 

\clearpage
\newpage
\bibliographystyle{assets/plainnat}
\bibliography{paper}

\begin{thebibliography}{115}
\providecommand{\natexlab}[1]{#1}
\providecommand{\url}[1]{\texttt{#1}}
\expandafter\ifx\csname urlstyle\endcsname\relax
  \providecommand{\doi}[1]{doi: #1}\else
  \providecommand{\doi}{doi: \begingroup \urlstyle{rm}\Url}\fi

\bibitem[Achanta et~al.(2012)Achanta, Shaji, Smith, Lucchi, Fua, and S{\"u}sstrunk]{achanta2012slic}
Radhakrishna Achanta, Appu Shaji, Kevin Smith, Aurelien Lucchi, Pascal Fua, and Sabine S{\"u}sstrunk.
\newblock Slic superpixels compared to state-of-the-art superpixel methods.
\newblock \emph{IEEE transactions on pattern analysis and machine intelligence}, 34\penalty0 (11):\penalty0 2274--2282, 2012.

\bibitem[Assran et~al.(2023)Assran, Duval, Misra, Bojanowski, Vincent, Rabbat, LeCun, and Ballas]{assran2023self}
Mahmoud Assran, Quentin Duval, Ishan Misra, Piotr Bojanowski, Pascal Vincent, Michael Rabbat, Yann LeCun, and Nicolas Ballas.
\newblock Self-supervised learning from images with a joint-embedding predictive architecture.
\newblock In \emph{Proceedings of the IEEE/CVF Conference on Computer Vision and Pattern Recognition}, pages 15619--15629, 2023.

\bibitem[Battaglia et~al.(2018)Battaglia, Hamrick, Bapst, Sanchez-Gonzalez, Zambaldi, Malinowski, Tacchetti, Raposo, Santoro, Faulkner, et~al.]{battaglia2018relationalinductivebiasesdeep}
Peter~W Battaglia, Jessica~B Hamrick, Victor Bapst, Alvaro Sanchez-Gonzalez, Vinicius Zambaldi, Mateusz Malinowski, Andrea Tacchetti, David Raposo, Adam Santoro, Ryan Faulkner, et~al.
\newblock Relational inductive biases, deep learning, and graph networks.
\newblock \emph{arXiv preprint arXiv:1806.01261}, 2018.

\bibitem[Behrouz and Hashemi(2024)]{behrouz2024graph}
Ali Behrouz and Farnoosh Hashemi.
\newblock Graph mamba: Towards learning on graphs with state space models.
\newblock In \emph{Proceedings of the 30th ACM SIGKDD Conference on Knowledge Discovery and Data Mining}, pages 119--130, 2024.

\bibitem[Bo et~al.(2023)Bo, Shi, Wang, and Liao]{bo2023specformer}
Deyu Bo, Chuan Shi, Lele Wang, and Renjie Liao.
\newblock Specformer: Spectral graph neural networks meet transformers.
\newblock In \emph{The Eleventh International Conference on Learning Representations}, 2023.

\bibitem[Bojchevski and G{\"u}nnemann(2017)]{bojchevski2017deep}
Aleksandar Bojchevski and Stephan G{\"u}nnemann.
\newblock Deep gaussian embedding of graphs: Unsupervised inductive learning via ranking.
\newblock \emph{arXiv preprint arXiv:1707.03815}, 2017.

\bibitem[Bojchevski et~al.(2020)Bojchevski, Gasteiger, Perozzi, Kapoor, Blais, R{\'o}zemberczki, Lukasik, and G{\"u}nnemann]{bojchevski2020scaling}
Aleksandar Bojchevski, Johannes Gasteiger, Bryan Perozzi, Amol Kapoor, Martin Blais, Benedek R{\'o}zemberczki, Michal Lukasik, and Stephan G{\"u}nnemann.
\newblock Scaling graph neural networks with approximate pagerank.
\newblock In \emph{Proceedings of the 26th ACM SIGKDD International Conference on Knowledge Discovery \& Data Mining}, pages 2464--2473, 2020.

\bibitem[Brown et~al.(2020)Brown, Mann, Ryder, Subbiah, Kaplan, Dhariwal, Neelakantan, Shyam, Sastry, Askell, Agarwal, Herbert-Voss, Krueger, Henighan, Child, Ramesh, Ziegler, Wu, Winter, Hesse, Chen, Sigler, Litwin, Gray, Chess, Clark, Berner, McCandlish, Radford, Sutskever, and Amodei]{NEURIPS2020_1457c0d6}
Tom Brown, Benjamin Mann, Nick Ryder, Melanie Subbiah, Jared~D Kaplan, Prafulla Dhariwal, Arvind Neelakantan, Pranav Shyam, Girish Sastry, Amanda Askell, Sandhini Agarwal, Ariel Herbert-Voss, Gretchen Krueger, Tom Henighan, Rewon Child, Aditya Ramesh, Daniel Ziegler, Jeffrey Wu, Clemens Winter, Chris Hesse, Mark Chen, Eric Sigler, Mateusz Litwin, Scott Gray, Benjamin Chess, Jack Clark, Christopher Berner, Sam McCandlish, Alec Radford, Ilya Sutskever, and Dario Amodei.
\newblock Language models are few-shot learners.
\newblock In \emph{Advances in Neural Information Processing Systems}, pages 1877--1901, 2020.

\bibitem[Cant\"{u}rk et~al.(2024)Cant\"{u}rk, Liu, Lapointe-Gagn\'{e}, L\'{e}tourneau, Wolf, Beaini, and Ramp\'{a}\v{s}ek]{pmlr-v235-canturk24a}
Semih Cant\"{u}rk, Renming Liu, Olivier Lapointe-Gagn\'{e}, Vincent L\'{e}tourneau, Guy Wolf, Dominique Beaini, and Ladislav Ramp\'{a}\v{s}ek.
\newblock Graph positional and structural encoder.
\newblock In \emph{Proceedings of the 41st International Conference on Machine Learning}, pages 5533--5566, 2024.

\bibitem[Chen et~al.(2022)Chen, O'Bray, and Borgwardt]{pmlr-v162-chen22r}
Dexiong Chen, Leslie O'Bray, and Karsten Borgwardt.
\newblock Structure-aware transformer for graph representation learning.
\newblock In \emph{Proceedings of the 39th International Conference on Machine Learning}, pages 3469--3489, 2022.

\bibitem[Chen et~al.(2023)Chen, Gao, Li, and He]{chen2023nagphormer}
Jinsong Chen, Kaiyuan Gao, Gaichao Li, and Kun He.
\newblock {NAG}phormer: A tokenized graph transformer for node classification in large graphs.
\newblock In \emph{The Eleventh International Conference on Learning Representations}, 2023.

\bibitem[Chien et~al.(2020)Chien, Peng, Li, and Milenkovic]{chien2020adaptive}
Eli Chien, Jianhao Peng, Pan Li, and Olgica Milenkovic.
\newblock Adaptive universal generalized pagerank graph neural network.
\newblock \emph{arXiv preprint arXiv:2006.07988}, 2020.

\bibitem[Dao et~al.(2022)Dao, Fu, Ermon, Rudra, and R{\'e}]{dao2022flashattention}
Tri Dao, Dan Fu, Stefano Ermon, Atri Rudra, and Christopher R{\'e}.
\newblock Flashattention: Fast and memory-efficient exact attention with io-awareness.
\newblock \emph{Advances in Neural Information Processing Systems}, 35:\penalty0 16344--16359, 2022.

\bibitem[Deng et~al.(2024)Deng, Yue, and Zhang]{deng2024polynormer}
Chenhui Deng, Zichao Yue, and Zhiru Zhang.
\newblock Polynormer: Polynomial-expressive graph transformer in linear time.
\newblock In \emph{The Twelfth International Conference on Learning Representations}, 2024.

\bibitem[Devlin et~al.(2019)Devlin, Chang, Lee, and Toutanova]{devlin-etal-2019-bert}
Jacob Devlin, Ming-Wei Chang, Kenton Lee, and Kristina Toutanova.
\newblock {BERT}: Pre-training of deep bidirectional transformers for language understanding.
\newblock In \emph{Proceedings of the 2019 Conference of the North {A}merican Chapter of the Association for Computational Linguistics: Human Language Technologies}, pages 4171--4186, June 2019.

\bibitem[Dhariwal et~al.(2020)Dhariwal, Jun, Payne, Kim, Radford, and Sutskever]{dhariwal2020jukebox}
Prafulla Dhariwal, Heewoo Jun, Christine Payne, Jong~Wook Kim, Alec Radford, and Ilya Sutskever.
\newblock Jukebox: A generative model for music.
\newblock \emph{arXiv preprint arXiv:2005.00341}, 2020.

\bibitem[Ding et~al.(2021)Ding, Kong, Li, Zhu, Dickerson, Huang, and Goldstein]{ding2021vqgnn}
Mucong Ding, Kezhi Kong, Jingling Li, Chen Zhu, John~P Dickerson, Furong Huang, and Tom Goldstein.
\newblock {VQ}-{GNN}: A universal framework to scale up graph neural networks using vector quantization.
\newblock In A.~Beygelzimer, Y.~Dauphin, P.~Liang, and J.~Wortman Vaughan, editors, \emph{Advances in Neural Information Processing Systems}, 2021.

\bibitem[Doersch and Zisserman(2017)]{doersch2017multi}
Carl Doersch and Andrew Zisserman.
\newblock Multi-task self-supervised visual learning.
\newblock In \emph{Proceedings of the IEEE international conference on computer vision}, pages 2051--2060, 2017.

\bibitem[Dosovitskiy et~al.(2021)Dosovitskiy, Beyer, Kolesnikov, Weissenborn, Zhai, Unterthiner, Dehghani, Minderer, Heigold, Gelly, Uszkoreit, and Houlsby]{dosovitskiy2021an}
Alexey Dosovitskiy, Lucas Beyer, Alexander Kolesnikov, Dirk Weissenborn, Xiaohua Zhai, Thomas Unterthiner, Mostafa Dehghani, Matthias Minderer, Georg Heigold, Sylvain Gelly, Jakob Uszkoreit, and Neil Houlsby.
\newblock An image is worth 16x16 words: Transformers for image recognition at scale.
\newblock In \emph{International Conference on Learning Representations}, 2021.

\bibitem[Dubey et~al.(2024)Dubey, Jauhri, Pandey, Kadian, Al-Dahle, Letman, Mathur, Schelten, Yang, Fan, et~al.]{dubey2024llama}
Abhimanyu Dubey, Abhinav Jauhri, Abhinav Pandey, Abhishek Kadian, Ahmad Al-Dahle, Aiesha Letman, Akhil Mathur, Alan Schelten, Amy Yang, Angela Fan, et~al.
\newblock The llama 3 herd of models.
\newblock \emph{arXiv preprint arXiv:2407.21783}, 2024.

\bibitem[Dwivedi and Bresson(2020)]{dwivedi2020generalization}
Vijay~Prakash Dwivedi and Xavier Bresson.
\newblock A generalization of transformer networks to graphs.
\newblock \emph{arXiv preprint arXiv:2012.09699}, 2020.

\bibitem[Dwivedi et~al.(2022{\natexlab{a}})Dwivedi, Luu, Laurent, Bengio, and Bresson]{dwivedi2022graph}
Vijay~Prakash Dwivedi, Anh~Tuan Luu, Thomas Laurent, Yoshua Bengio, and Xavier Bresson.
\newblock Graph neural networks with learnable structural and positional representations.
\newblock In \emph{International Conference on Learning Representations}, 2022{\natexlab{a}}.

\bibitem[Dwivedi et~al.(2022{\natexlab{b}})Dwivedi, Ramp{\'a}{\v{s}}ek, Galkin, Parviz, Wolf, Luu, and Beaini]{dwivedi2022long}
Vijay~Prakash Dwivedi, Ladislav Ramp{\'a}{\v{s}}ek, Michael Galkin, Ali Parviz, Guy Wolf, Anh~Tuan Luu, and Dominique Beaini.
\newblock Long range graph benchmark.
\newblock \emph{Advances in Neural Information Processing Systems}, 35:\penalty0 22326--22340, 2022{\natexlab{b}}.

\bibitem[Fatemi et~al.(2024)Fatemi, Halcrow, and Perozzi]{fatemi2024talk}
Bahare Fatemi, Jonathan Halcrow, and Bryan Perozzi.
\newblock Talk like a graph: Encoding graphs for large language models.
\newblock In \emph{The Twelfth International Conference on Learning Representations}, 2024.

\bibitem[Feng et~al.(2022)Feng, Dong, Huang, Yin, Cheng, Kharlamov, and Tang]{feng2022grand+}
Wenzheng Feng, Yuxiao Dong, Tinglin Huang, Ziqi Yin, Xu~Cheng, Evgeny Kharlamov, and Jie Tang.
\newblock Grand+: Scalable graph random neural networks.
\newblock In \emph{Proceedings of the ACM Web Conference 2022}, pages 3248--3258, 2022.

\bibitem[Frasca et~al.(2020)Frasca, Rossi, Eynard, Chamberlain, Bronstein, and Monti]{frasca2020sign}
Fabrizio Frasca, Emanuele Rossi, Davide Eynard, Ben Chamberlain, Michael Bronstein, and Federico Monti.
\newblock Sign: Scalable inception graph neural networks.
\newblock \emph{arXiv preprint arXiv:2004.11198}, 2020.

\bibitem[Fu et~al.(2020)Fu, Xu, Li, Tong, and He]{DBLP:conf/cikm/FuXLTH20}
Dongqi Fu, Zhe Xu, Bo~Li, Hanghang Tong, and Jingrui He.
\newblock A view-adversarial framework for multi-view network embedding.
\newblock In \emph{CIKM}, 2020.

\bibitem[Fu et~al.(2024)Fu, Hua, Xie, Fang, Zhang, Sancak, Wu, Malevich, He, and Long]{fu2024vcrgraphormer}
Dongqi Fu, Zhigang Hua, Yan Xie, Jin Fang, Si~Zhang, Kaan Sancak, Hao Wu, Andrey Malevich, Jingrui He, and Bo~Long.
\newblock {VCR}-graphormer: A mini-batch graph transformer via virtual connections.
\newblock In \emph{The Twelfth International Conference on Learning Representations}, 2024.

\bibitem[Galkin et~al.(2022)Galkin, Denis, Wu, and Hamilton]{galkin2022nodepiece}
Mikhail Galkin, Etienne Denis, Jiapeng Wu, and William~L. Hamilton.
\newblock Nodepiece: Compositional and parameter-efficient representations of large knowledge graphs.
\newblock In \emph{International Conference on Learning Representations}, 2022.

\bibitem[Gasteiger et~al.(2018)Gasteiger, Bojchevski, and G{\"u}nnemann]{gasteiger2018predict}
Johannes Gasteiger, Aleksandar Bojchevski, and Stephan G{\"u}nnemann.
\newblock Predict then propagate: Graph neural networks meet personalized pagerank.
\newblock \emph{arXiv preprint arXiv:1810.05997}, 2018.

\bibitem[Geisler et~al.(2021)Geisler, Schmidt, {\c{S}}irin, Z{\"u}gner, Bojchevski, and G{\"u}nnemann]{geisler2021robustness}
Simon Geisler, Tobias Schmidt, Hakan {\c{S}}irin, Daniel Z{\"u}gner, Aleksandar Bojchevski, and Stephan G{\"u}nnemann.
\newblock Robustness of graph neural networks at scale.
\newblock \emph{Advances in Neural Information Processing Systems}, 34:\penalty0 7637--7649, 2021.

\bibitem[Ghiasi et~al.(2021)Ghiasi, Zoph, Cubuk, Le, and Lin]{ghiasi2021multi}
Golnaz Ghiasi, Barret Zoph, Ekin~D Cubuk, Quoc~V Le, and Tsung-Yi Lin.
\newblock Multi-task self-training for learning general representations.
\newblock In \emph{Proceedings of the IEEE/CVF International Conference on Computer Vision}, pages 8856--8865, 2021.

\bibitem[Gilmer et~al.(2017)Gilmer, Schoenholz, Riley, Vinyals, and Dahl]{gilmer2017neural}
Justin Gilmer, Samuel~S Schoenholz, Patrick~F Riley, Oriol Vinyals, and George~E Dahl.
\newblock Neural message passing for quantum chemistry.
\newblock In \emph{International conference on machine learning}, pages 1263--1272, 2017.

\bibitem[Hamilton et~al.(2017)Hamilton, Ying, and Leskovec]{hamilton2017inductive}
Will Hamilton, Zhitao Ying, and Jure Leskovec.
\newblock Inductive representation learning on large graphs.
\newblock \emph{Advances in neural information processing systems}, 30, 2017.

\bibitem[Hassani and Khasahmadi(2020)]{pmlr-v119-hassani20a}
Kaveh Hassani and Amir~Hosein Khasahmadi.
\newblock Contrastive multi-view representation learning on graphs.
\newblock In \emph{International Conference on Machine Learning}, pages 4116--4126, 2020.

\bibitem[He et~al.(2024)He, Bresson, Laurent, Perold, LeCun, and Hooi]{he2024harnessing}
Xiaoxin He, Xavier Bresson, Thomas Laurent, Adam Perold, Yann LeCun, and Bryan Hooi.
\newblock Harnessing explanations: {LLM}-to-{LM} interpreter for enhanced text-attributed graph representation learning.
\newblock In \emph{The Twelfth International Conference on Learning Representations}, 2024.

\bibitem[Hoang et~al.(2024)Hoang, Lee, et~al.]{hoang2024survey}
Van~Thuy Hoang, O~Lee, et~al.
\newblock A survey on structure-preserving graph transformers.
\newblock \emph{arXiv preprint arXiv:2401.16176}, 2024.

\bibitem[Hou et~al.(2020)Hou, Zhang, Cheng, Ma, Ma, Chen, and Yang]{Hou2020Measuring}
Yifan Hou, Jian Zhang, James Cheng, Kaili Ma, Richard T.~B. Ma, Hongzhi Chen, and Ming-Chang Yang.
\newblock Measuring and improving the use of graph information in graph neural networks.
\newblock In \emph{International Conference on Learning Representations}, 2020.

\bibitem[Hou et~al.(2022)Hou, Liu, Cen, Dong, Yang, Wang, and Tang]{hou2022graphmae}
Zhenyu Hou, Xiao Liu, Yukuo Cen, Yuxiao Dong, Hongxia Yang, Chunjie Wang, and Jie Tang.
\newblock Graphmae: Self-supervised masked graph autoencoders.
\newblock In \emph{Proceedings of the 28th ACM SIGKDD Conference on Knowledge Discovery and Data Mining}, pages 594--604, 2022.

\bibitem[Hou et~al.(2023)Hou, He, Cen, Liu, Dong, Kharlamov, and Tang]{hou2023graphmae2}
Zhenyu Hou, Yufei He, Yukuo Cen, Xiao Liu, Yuxiao Dong, Evgeny Kharlamov, and Jie Tang.
\newblock Graphmae2: A decoding-enhanced masked self-supervised graph learner.
\newblock In \emph{Proceedings of the ACM web conference 2023}, pages 737--746, 2023.

\bibitem[Hu et~al.(2020{\natexlab{a}})Hu, Fey, Zitnik, Dong, Ren, Liu, Catasta, and Leskovec]{hu2020open}
Weihua Hu, Matthias Fey, Marinka Zitnik, Yuxiao Dong, Hongyu Ren, Bowen Liu, Michele Catasta, and Jure Leskovec.
\newblock Open graph benchmark: Datasets for machine learning on graphs.
\newblock \emph{Advances in neural information processing systems}, 33:\penalty0 22118--22133, 2020{\natexlab{a}}.

\bibitem[Hu et~al.(2020{\natexlab{b}})Hu, Liu, Gomes, Zitnik, Liang, Pande, and Leskovec]{Hu2020Strategies}
Weihua Hu, Bowen Liu, Joseph Gomes, Marinka Zitnik, Percy Liang, Vijay Pande, and Jure Leskovec.
\newblock Strategies for pre-training graph neural networks.
\newblock In \emph{International Conference on Learning Representations}, 2020{\natexlab{b}}.

\bibitem[Jin et~al.(2018)Jin, Barzilay, and Jaakkola]{jin2018junction}
Wengong Jin, Regina Barzilay, and Tommi Jaakkola.
\newblock Junction tree variational autoencoder for molecular graph generation.
\newblock In \emph{International conference on machine learning}, pages 2323--2332, 2018.

\bibitem[Khasahmadi et~al.(2020)Khasahmadi, Hassani, Moradi, Lee, and Morris]{Khasahmadi2020Memory-Based}
Amir~Hosein Khasahmadi, Kaveh Hassani, Parsa Moradi, Leo Lee, and Quaid Morris.
\newblock Memory-based graph networks.
\newblock In \emph{International Conference on Learning Representations}, 2020.

\bibitem[Kim et~al.(2022)Kim, Nguyen, Min, Cho, Lee, Lee, and Hong]{kim2022pure}
Jinwoo Kim, Dat Nguyen, Seonwoo Min, Sungjun Cho, Moontae Lee, Honglak Lee, and Seunghoon Hong.
\newblock Pure transformers are powerful graph learners.
\newblock \emph{Advances in Neural Information Processing Systems}, pages 14582--14595, 2022.

\bibitem[Kipf and Welling(2017)]{kipf2017semisupervised}
Thomas~N. Kipf and Max Welling.
\newblock Semi-supervised classification with graph convolutional networks.
\newblock In \emph{International Conference on Learning Representations}, 2017.

\bibitem[Kong et~al.(2023)Kong, Chen, Kirchenbauer, Ni, Bruss, and Goldstein]{kong2023goat}
Kezhi Kong, Jiuhai Chen, John Kirchenbauer, Renkun Ni, C~Bayan Bruss, and Tom Goldstein.
\newblock Goat: A global transformer on large-scale graphs.
\newblock In \emph{International Conference on Machine Learning}, pages 17375--17390, 2023.

\bibitem[Kreuzer et~al.(2021)Kreuzer, Beaini, Hamilton, L{\'e}tourneau, and Tossou]{kreuzer2021rethinking}
Devin Kreuzer, Dominique Beaini, Will Hamilton, Vincent L{\'e}tourneau, and Prudencio Tossou.
\newblock Rethinking graph transformers with spectral attention.
\newblock \emph{Advances in Neural Information Processing Systems}, 34:\penalty0 21618--21629, 2021.

\bibitem[Lee et~al.(2022)Lee, Kim, Kim, Cho, and Han]{lee2022autoregressive}
Doyup Lee, Chiheon Kim, Saehoon Kim, Minsu Cho, and Wook-Shin Han.
\newblock Autoregressive image generation using residual quantization.
\newblock In \emph{Proceedings of the IEEE/CVF Conference on Computer Vision and Pattern Recognition}, pages 11523--11532, 2022.

\bibitem[Leskovec and Krevl(2014)]{leskovec2016snap}
Jure Leskovec and Andrej Krevl.
\newblock {SNAP Datasets}: {Stanford} large network dataset collection.
\newblock \url{http://snap.stanford.edu/data}, June 2014.

\bibitem[Li et~al.(2022)Li, Zhu, Cheng, Shan, Luo, Li, and Qian]{li2022finding}
Xiang Li, Renyu Zhu, Yao Cheng, Caihua Shan, Siqiang Luo, Dongsheng Li, and Weining Qian.
\newblock Finding global homophily in graph neural networks when meeting heterophily.
\newblock In \emph{International Conference on Machine Learning}, pages 13242--13256, 2022.

\bibitem[Li et~al.(2024)Li, Wang, Luo, Edwards, Gui, Lin, Ji, and Ji]{li2024geometry}
Xiner Li, Limei Wang, Youzhi Luo, Carl Edwards, Shurui Gui, Yuchao Lin, Heng Ji, and Shuiwang Ji.
\newblock Geometry informed tokenization of molecules for language model generation.
\newblock \emph{arXiv preprint arXiv:2408.10120}, 2024.

\bibitem[Lim et~al.(2021)Lim, Hohne, Li, Huang, Gupta, Bhalerao, and Lim]{lim2021large}
Derek Lim, Felix Hohne, Xiuyu Li, Sijia~Linda Huang, Vaishnavi Gupta, Omkar Bhalerao, and Ser~Nam Lim.
\newblock Large scale learning on non-homophilous graphs: New benchmarks and strong simple methods.
\newblock \emph{Advances in Neural Information Processing Systems}, 34:\penalty0 20887--20902, 2021.

\bibitem[Lin et~al.(2014)Lin, Maire, Belongie, Hays, Perona, Ramanan, Doll{\'a}r, and Zitnick]{lin2014microsoft}
Tsung-Yi Lin, Michael Maire, Serge Belongie, James Hays, Pietro Perona, Deva Ramanan, Piotr Doll{\'a}r, and C~Lawrence Zitnick.
\newblock Microsoft coco: Common objects in context.
\newblock In \emph{Computer vision--ECCV 2014: 13th European conference, zurich, Switzerland, September 6-12, 2014, proceedings, part v 13}, pages 740--755. Springer, 2014.

\bibitem[Liu et~al.(2023{\natexlab{a}})Liu, Zhan, Ma, Ding, Tao, Wu, and Hu]{10.24963/ijcai.2023/244}
Chuang Liu, Yibing Zhan, Xueqi Ma, Liang Ding, Dapeng Tao, Jia Wu, and Wenbin Hu.
\newblock Gapformer: graph transformer with graph pooling for node classification.
\newblock In \emph{Proceedings of the Thirty-Second International Joint Conference on Artificial Intelligence}, 2023{\natexlab{a}}.

\bibitem[Liu et~al.(2024)Liu, Zaharia, and Abbeel]{liu2024ringattention}
Hao Liu, Matei Zaharia, and Pieter Abbeel.
\newblock Ringattention with blockwise transformers for near-infinite context.
\newblock In \emph{The Twelfth International Conference on Learning Representations}, 2024.

\bibitem[Liu et~al.(2022)Liu, HaoChen, Gaidon, and Ma]{liu2022selfsupervised}
Hong Liu, Jeff~Z. HaoChen, Adrien Gaidon, and Tengyu Ma.
\newblock Self-supervised learning is more robust to dataset imbalance.
\newblock In \emph{International Conference on Learning Representations}, 2022.

\bibitem[Liu et~al.(2023{\natexlab{b}})Liu, Yang, Lu, Chen, Li, Zhang, Bai, Fang, Sun, Yu, et~al.]{liu2023towards}
Jiawei Liu, Cheng Yang, Zhiyuan Lu, Junze Chen, Yibo Li, Mengmei Zhang, Ting Bai, Yuan Fang, Lichao Sun, Philip~S Yu, et~al.
\newblock Towards graph foundation models: A survey and beyond.
\newblock \emph{arXiv preprint arXiv:2310.11829}, 2023{\natexlab{b}}.

\bibitem[Liu et~al.(2021)Liu, Lin, Cao, Hu, Wei, Zhang, Lin, and Guo]{liu2021swin}
Ze~Liu, Yutong Lin, Yue Cao, Han Hu, Yixuan Wei, Zheng Zhang, Stephen Lin, and Baining Guo.
\newblock Swin transformer: Hierarchical vision transformer using shifted windows.
\newblock In \emph{Proceedings of the IEEE/CVF international conference on computer vision}, pages 10012--10022, 2021.

\bibitem[Liu et~al.(2023{\natexlab{c}})Liu, Shi, Zhang, Zhang, Kawaguchi, Wang, and Chua]{liu2023rethinking}
Zhiyuan Liu, Yaorui Shi, An~Zhang, Enzhi Zhang, Kenji Kawaguchi, Xiang Wang, and Tat-Seng Chua.
\newblock Rethinking tokenizer and decoder in masked graph modeling for molecules.
\newblock In \emph{Thirty-seventh Conference on Neural Information Processing Systems}, 2023{\natexlab{c}}.

\bibitem[Luo et~al.(2024{\natexlab{a}})Luo, Liu, Shi, and Wu]{luo2024structure}
Yuankai Luo, Qijiong Liu, Lei Shi, and Xiao-Ming Wu.
\newblock Structure-aware semantic node identifiers for learning on graphs.
\newblock \emph{arXiv preprint arXiv:2405.16435}, 2024{\natexlab{a}}.

\bibitem[Luo et~al.(2024{\natexlab{b}})Luo, Shi, and Wu]{luo2024classic}
Yuankai Luo, Lei Shi, and Xiao-Ming Wu.
\newblock Classic {GNN}s are strong baselines: Reassessing {GNN}s for node classification.
\newblock In \emph{The Thirty-eight Conference on Neural Information Processing Systems Datasets and Benchmarks Track}, 2024{\natexlab{b}}.

\bibitem[Ma et~al.(2023)Ma, Lin, Lim, Romero-Soriano, Dokania, Coates, Torr, and Lim]{pmlr-v202-ma23c}
Liheng Ma, Chen Lin, Derek Lim, Adriana Romero-Soriano, Puneet~K. Dokania, Mark Coates, Philip Torr, and Ser-Nam Lim.
\newblock Graph inductive biases in transformers without message passing.
\newblock In \emph{Proceedings of the 40th International Conference on Machine Learning}, pages 23321--23337, 2023.

\bibitem[Mao et~al.(2024)Mao, Chen, Tang, Zhao, Ma, Zhao, Shah, Galkin, and Tang]{mao2024position}
Haitao Mao, Zhikai Chen, Wenzhuo Tang, Jianan Zhao, Yao Ma, Tong Zhao, Neil Shah, Mikhail Galkin, and Jiliang Tang.
\newblock Position: Graph foundation models are already here.
\newblock In \emph{Forty-first International Conference on Machine Learning}, 2024.

\bibitem[McAuley et~al.(2015)McAuley, Targett, Shi, and Van Den~Hengel]{mcauley2015image}
Julian McAuley, Christopher Targett, Qinfeng Shi, and Anton Van Den~Hengel.
\newblock Image-based recommendations on styles and substitutes.
\newblock In \emph{Proceedings of the 38th international ACM SIGIR conference on research and development in information retrieval}, pages 43--52, 2015.

\bibitem[Mernyei and Cangea(2020)]{mernyei2020wiki}
P{\'e}ter Mernyei and C{\u{a}}t{\u{a}}lina Cangea.
\newblock Wiki-cs: A wikipedia-based benchmark for graph neural networks.
\newblock \emph{arXiv preprint arXiv:2007.02901}, 2020.

\bibitem[Mialon et~al.(2021)Mialon, Chen, Selosse, and Mairal]{mialon2021graphit}
Gr{\'e}goire Mialon, Dexiong Chen, Margot Selosse, and Julien Mairal.
\newblock Graphit: Encoding graph structure in transformers.
\newblock \emph{arXiv preprint arXiv:2106.05667}, 2021.

\bibitem[M{\"u}ller et~al.(2024)M{\"u}ller, Galkin, Morris, and Ramp{\'a}{\v{s}}ek]{muller2024attending}
Luis M{\"u}ller, Mikhail Galkin, Christopher Morris, and Ladislav Ramp{\'a}{\v{s}}ek.
\newblock Attending to graph transformers.
\newblock \emph{Transactions on Machine Learning Research}, 2024.
\newblock ISSN 2835-8856.

\bibitem[Nakata and Shimazaki(2017)]{nakata2017pubchemqc}
Maho Nakata and Tomomi Shimazaki.
\newblock Pubchemqc project: a large-scale first-principles electronic structure database for data-driven chemistry.
\newblock \emph{Journal of chemical information and modeling}, 57\penalty0 (6):\penalty0 1300--1308, 2017.

\bibitem[Namata et~al.(2012)Namata, London, Getoor, Huang, and Edu]{namata2012query}
Galileo Namata, Ben London, Lise Getoor, Bert Huang, and U~Edu.
\newblock Query-driven active surveying for collective classification.
\newblock In \emph{10th international workshop on mining and learning with graphs}, volume~8, page~1, 2012.

\bibitem[Pei et~al.(2020)Pei, Wei, Chang, Lei, and Yang]{pei2020geom}
Hongbin Pei, Bingzhe Wei, Kevin Chen-Chuan Chang, Yu~Lei, and Bo~Yang.
\newblock Geom-gcn: Geometric graph convolutional networks.
\newblock \emph{arXiv preprint arXiv:2002.05287}, 2020.

\bibitem[Platonov et~al.(2023)Platonov, Kuznedelev, Diskin, Babenko, and Prokhorenkova]{platonov2023critical}
Oleg Platonov, Denis Kuznedelev, Michael Diskin, Artem Babenko, and Liudmila Prokhorenkova.
\newblock A critical look at the evaluation of {GNN}s under heterophily: Are we really making progress?
\newblock In \emph{The Eleventh International Conference on Learning Representations}, 2023.

\bibitem[Ramesh et~al.(2021)Ramesh, Pavlov, Goh, Gray, Voss, Radford, Chen, and Sutskever]{pmlr-v139-ramesh21a}
Aditya Ramesh, Mikhail Pavlov, Gabriel Goh, Scott Gray, Chelsea Voss, Alec Radford, Mark Chen, and Ilya Sutskever.
\newblock Zero-shot text-to-image generation.
\newblock In \emph{Proceedings of the 38th International Conference on Machine Learning}, pages 8821--8831, 2021.

\bibitem[Ramp{\'a}{\v{s}}ek et~al.(2022)Ramp{\'a}{\v{s}}ek, Galkin, Dwivedi, Luu, Wolf, and Beaini]{rampavsek2022recipe}
Ladislav Ramp{\'a}{\v{s}}ek, Michael Galkin, Vijay~Prakash Dwivedi, Anh~Tuan Luu, Guy Wolf, and Dominique Beaini.
\newblock Recipe for a general, powerful, scalable graph transformer.
\newblock \emph{Advances in Neural Information Processing Systems}, 35:\penalty0 14501--14515, 2022.

\bibitem[Reid et~al.(2024)Reid, Savinov, Teplyashin, Lepikhin, Lillicrap, Alayrac, Soricut, Lazaridou, Firat, Schrittwieser, et~al.]{reid2024gemini}
Machel Reid, Nikolay Savinov, Denis Teplyashin, Dmitry Lepikhin, Timothy Lillicrap, Jean-baptiste Alayrac, Radu Soricut, Angeliki Lazaridou, Orhan Firat, Julian Schrittwieser, et~al.
\newblock Gemini 1.5: Unlocking multimodal understanding across millions of tokens of context.
\newblock \emph{arXiv preprint arXiv:2403.05530}, 2024.

\bibitem[Rong et~al.(2020)Rong, Bian, Xu, Xie, Wei, Huang, and Huang]{rong2020self}
Yu~Rong, Yatao Bian, Tingyang Xu, Weiyang Xie, Ying Wei, Wenbing Huang, and Junzhou Huang.
\newblock Self-supervised graph transformer on large-scale molecular data.
\newblock \emph{Advances in neural information processing systems}, pages 12559--12571, 2020.

\bibitem[Rozemberczki et~al.(2021)Rozemberczki, Allen, and Sarkar]{rozemberczki2021multi}
Benedek Rozemberczki, Carl Allen, and Rik Sarkar.
\newblock Multi-scale attributed node embedding.
\newblock \emph{Journal of Complex Networks}, 9\penalty0 (2):\penalty0 cnab014, 2021.

\bibitem[Shchur et~al.(2018)Shchur, Mumme, Bojchevski, and G{\"u}nnemann]{shchur2018pitfalls}
Oleksandr Shchur, Maximilian Mumme, Aleksandar Bojchevski, and Stephan G{\"u}nnemann.
\newblock Pitfalls of graph neural network evaluation.
\newblock \emph{arXiv preprint arXiv:1811.05868}, 2018.

\bibitem[Shi et~al.(2023)Shi, Daunhawer, Vogt, Torr, and Sanyal]{shi2023how}
Yuge Shi, Imant Daunhawer, Julia~E Vogt, Philip Torr, and Amartya Sanyal.
\newblock How robust is unsupervised representation learning to distribution shift?
\newblock In \emph{The Eleventh International Conference on Learning Representations}, 2023.

\bibitem[Shirzad et~al.(2023)Shirzad, Velingker, Venkatachalam, Sutherland, and Sinop]{pmlr-v202-shirzad23a}
Hamed Shirzad, Ameya Velingker, Balaji Venkatachalam, Danica~J. Sutherland, and Ali~Kemal Sinop.
\newblock Exphormer: Sparse transformers for graphs.
\newblock In \emph{Proceedings of the 40th International Conference on Machine Learning}, pages 31613--31632, 2023.

\bibitem[Singh et~al.(2016)Singh, Chaudhary, Dhanda, Bhalla, Usmani, Gautam, Tuknait, Agrawal, Mathur, and Raghava]{singh2016satpdb}
Sandeep Singh, Kumardeep Chaudhary, Sandeep~Kumar Dhanda, Sherry Bhalla, Salman~Sadullah Usmani, Ankur Gautam, Abhishek Tuknait, Piyush Agrawal, Deepika Mathur, and Gajendra~PS Raghava.
\newblock Satpdb: a database of structurally annotated therapeutic peptides.
\newblock \emph{Nucleic acids research}, 44\penalty0 (D1):\penalty0 D1119--D1126, 2016.

\bibitem[Sun et~al.(2020)Sun, Hoffman, Verma, and Tang]{Sun2020InfoGraph}
Fan-Yun Sun, Jordan Hoffman, Vikas Verma, and Jian Tang.
\newblock Infograph: Unsupervised graph-level representation learning via mutual information maximization.
\newblock In \emph{International Conference on Learning Representations}, 2020.

\bibitem[Team(2024)]{team2024chameleon}
Chameleon Team.
\newblock Chameleon: Mixed-modal early-fusion foundation models.
\newblock \emph{arXiv preprint arXiv:2405.09818}, 2024.

\bibitem[Thakoor et~al.(2022)Thakoor, Tallec, Azar, Azabou, Dyer, Munos, Veli{\v{c}}kovi{\'c}, and Valko]{thakoor2022largescale}
Shantanu Thakoor, Corentin Tallec, Mohammad~Gheshlaghi Azar, Mehdi Azabou, Eva~L Dyer, Remi Munos, Petar Veli{\v{c}}kovi{\'c}, and Michal Valko.
\newblock Large-scale representation learning on graphs via bootstrapping.
\newblock In \emph{International Conference on Learning Representations}, 2022.

\bibitem[Van Den~Oord et~al.(2017)Van Den~Oord, Vinyals, et~al.]{van2017neural}
Aaron Van Den~Oord, Oriol Vinyals, et~al.
\newblock Neural discrete representation learning.
\newblock \emph{Advances in neural information processing systems}, 30, 2017.

\bibitem[Van~Kempen et~al.(2024)Van~Kempen, Kim, Tumescheit, Mirdita, Lee, Gilchrist, S{\"o}ding, and Steinegger]{van2024fast}
Michel Van~Kempen, Stephanie~S Kim, Charlotte Tumescheit, Milot Mirdita, Jeongjae Lee, Cameron~LM Gilchrist, Johannes S{\"o}ding, and Martin Steinegger.
\newblock Fast and accurate protein structure search with foldseek.
\newblock \emph{Nature biotechnology}, 42\penalty0 (2):\penalty0 243--246, 2024.

\bibitem[Vaswani et~al.(2017)Vaswani, Shazeer, Parmar, Uszkoreit, Jones, Gomez, Kaiser, and Polosukhin]{NIPS2017_3f5ee243}
Ashish Vaswani, Noam Shazeer, Niki Parmar, Jakob Uszkoreit, Llion Jones, Aidan~N Gomez, \L~ukasz Kaiser, and Illia Polosukhin.
\newblock Attention is all you need.
\newblock In \emph{Advances in Neural Information Processing Systems}, 2017.

\bibitem[Veličković et~al.(2018)Veličković, Cucurull, Casanova, Romero, Liò, and Bengio]{veličković2018graph}
Petar Veličković, Guillem Cucurull, Arantxa Casanova, Adriana Romero, Pietro Liò, and Yoshua Bengio.
\newblock Graph attention networks.
\newblock In \emph{International Conference on Learning Representations}, 2018.

\bibitem[Veličković et~al.(2019)Veličković, Fedus, Hamilton, Liò, Bengio, and Hjelm]{velickovic_2019_iclr}
Petar Veličković, William Fedus, William~L. Hamilton, Pietro Liò, Yoshua Bengio, and R~Devon Hjelm.
\newblock Deep graph infomax.
\newblock In \emph{International Conference on Learning Representations}, 2019.

\bibitem[Wang et~al.(2024)Wang, Tsepa, Ma, and Wang]{wang2024graph}
Chloe Wang, Oleksii Tsepa, Jun Ma, and Bo~Wang.
\newblock Graph-mamba: Towards long-range graph sequence modeling with selective state spaces.
\newblock \emph{arXiv preprint arXiv:2402.00789}, 2024.

\bibitem[Wang et~al.(2020)Wang, Shen, Huang, Wu, Dong, and Kanakia]{wang2020microsoft}
Kuansan Wang, Zhihong Shen, Chiyuan Huang, Chieh-Han Wu, Yuxiao Dong, and Anshul Kanakia.
\newblock Microsoft academic graph: When experts are not enough.
\newblock \emph{Quantitative Science Studies}, 1\penalty0 (1):\penalty0 396--413, 2020.

\bibitem[Wu et~al.(2022)Wu, Zhao, Li, Wipf, and Yan]{wu2022nodeformer}
Qitian Wu, Wentao Zhao, Zenan Li, David Wipf, and Junchi Yan.
\newblock Nodeformer: A scalable graph structure learning transformer for node classification.
\newblock In Alice~H. Oh, Alekh Agarwal, Danielle Belgrave, and Kyunghyun Cho, editors, \emph{Advances in Neural Information Processing Systems}, 2022.

\bibitem[Wu et~al.(2023{\natexlab{a}})Wu, Yang, Zhao, He, Wipf, and Yan]{wu2023difformer}
Qitian Wu, Chenxiao Yang, Wentao Zhao, Yixuan He, David Wipf, and Junchi Yan.
\newblock {DIFF}ormer: Scalable (graph) transformers induced by energy constrained diffusion.
\newblock In \emph{The Eleventh International Conference on Learning Representations}, 2023{\natexlab{a}}.

\bibitem[Wu et~al.(2023{\natexlab{b}})Wu, Zhao, Yang, Zhang, Nie, Jiang, Bian, and Yan]{wu2023simplifying}
Qitian Wu, Wentao Zhao, Chenxiao Yang, Hengrui Zhang, Fan Nie, Haitian Jiang, Yatao Bian, and Junchi Yan.
\newblock Simplifying and empowering transformers for large-graph representations.
\newblock In \emph{Thirty-seventh Conference on Neural Information Processing Systems}, 2023{\natexlab{b}}.

\bibitem[Wu et~al.(2021)Wu, Jain, Wright, Mirhoseini, Gonzalez, and Stoica]{wu2021representing}
Zhanghao Wu, Paras Jain, Matthew Wright, Azalia Mirhoseini, Joseph~E Gonzalez, and Ion Stoica.
\newblock Representing long-range context for graph neural networks with global attention.
\newblock \emph{Advances in Neural Information Processing Systems}, pages 13266--13279, 2021.

\bibitem[Xia et~al.(2023)Xia, Zhao, Hu, Gao, Tan, Liu, Li, and Li]{xia2023molebert}
Jun Xia, Chengshuai Zhao, Bozhen Hu, Zhangyang Gao, Cheng Tan, Yue Liu, Siyuan Li, and Stan~Z. Li.
\newblock Mole-{BERT}: Rethinking pre-training graph neural networks for molecules.
\newblock In \emph{The Eleventh International Conference on Learning Representations}, 2023.

\bibitem[Xu et~al.(2025)Xu, Hassani, Zhang, Zeng, Yasunaga, Wang, Fu, Yao, Long, and Tong]{xu2025makellmsstrongnode}
Zhe Xu, Kaveh Hassani, Si~Zhang, Hanqing Zeng, Michihiro Yasunaga, Limei Wang, Dongqi Fu, Ning Yao, Bo~Long, and Hanghang Tong.
\newblock How to make llms strong node classifiers?
\newblock \emph{arXiv preprint arXiv:2410.02296}, 2025.

\bibitem[Yang et~al.(2024)Yang, Tian, Xu, Liu, Hong, Qu, Zhang, CUI, Zhang, and Leskovec]{yang2024vqgraph}
Ling Yang, Ye~Tian, Minkai Xu, Zhongyi Liu, Shenda Hong, Wei Qu, Wentao Zhang, Bin CUI, Muhan Zhang, and Jure Leskovec.
\newblock {VQG}raph: Rethinking graph representation space for bridging {GNN}s and {MLP}s.
\newblock In \emph{The Twelfth International Conference on Learning Representations}, 2024.

\bibitem[Yang et~al.(2016)Yang, Cohen, and Salakhudinov]{yang2016revisiting}
Zhilin Yang, William Cohen, and Ruslan Salakhudinov.
\newblock Revisiting semi-supervised learning with graph embeddings.
\newblock In \emph{International conference on machine learning}, pages 40--48. PMLR, 2016.

\bibitem[Ye et~al.(2024)Ye, Zhang, Wang, Xu, and Zhang]{ye-etal-2024-language}
Ruosong Ye, Caiqi Zhang, Runhui Wang, Shuyuan Xu, and Yongfeng Zhang.
\newblock Language is all a graph needs.
\newblock In \emph{Findings of the Association for Computational Linguistics: EACL 2024}, pages 1955--1973, 2024.

\bibitem[Ying et~al.(2021)Ying, Cai, Luo, Zheng, Ke, He, Shen, and Liu]{ying2021transformers}
Chengxuan Ying, Tianle Cai, Shengjie Luo, Shuxin Zheng, Guolin Ke, Di~He, Yanming Shen, and Tie-Yan Liu.
\newblock Do transformers really perform badly for graph representation?
\newblock \emph{Advances in neural information processing systems}, 34:\penalty0 28877--28888, 2021.

\bibitem[You et~al.(2020)You, Chen, Sui, Chen, Wang, and Shen]{you2020graph}
Yuning You, Tianlong Chen, Yongduo Sui, Ting Chen, Zhangyang Wang, and Yang Shen.
\newblock Graph contrastive learning with augmentations.
\newblock \emph{Advances in Neural Information Processing Systems}, 2020.

\bibitem[You et~al.(2021)You, Chen, Shen, and Wang]{you2021graph}
Yuning You, Tianlong Chen, Yang Shen, and Zhangyang Wang.
\newblock Graph contrastive learning automated.
\newblock In \emph{International Conference on Machine Learning}, pages 12121--12132, 2021.

\bibitem[Yu et~al.(2022{\natexlab{a}})Yu, Li, Koh, Zhang, Pang, Qin, Ku, Xu, Baldridge, and Wu]{yu2022vectorquantized}
Jiahui Yu, Xin Li, Jing~Yu Koh, Han Zhang, Ruoming Pang, James Qin, Alexander Ku, Yuanzhong Xu, Jason Baldridge, and Yonghui Wu.
\newblock Vector-quantized image modeling with improved {VQGAN}.
\newblock In \emph{International Conference on Learning Representations}, 2022{\natexlab{a}}.

\bibitem[Yu et~al.(2022{\natexlab{b}})Yu, Tang, Rao, Huang, Zhou, and Lu]{yu2022point}
Xumin Yu, Lulu Tang, Yongming Rao, Tiejun Huang, Jie Zhou, and Jiwen Lu.
\newblock Point-bert: Pre-training 3d point cloud transformers with masked point modeling.
\newblock In \emph{Proceedings of the IEEE/CVF conference on computer vision and pattern recognition}, pages 19313--19322, 2022{\natexlab{b}}.

\bibitem[Yuan et~al.(2021)Yuan, Chen, Wang, Yu, Shi, Jiang, Tay, Feng, and Yan]{Yuan_2021_ICCV}
Li~Yuan, Yunpeng Chen, Tao Wang, Weihao Yu, Yujun Shi, Zi-Hang Jiang, Francis~E.H. Tay, Jiashi Feng, and Shuicheng Yan.
\newblock Tokens-to-token vit: Training vision transformers from scratch on imagenet.
\newblock In \emph{Proceedings of the IEEE/CVF International Conference on Computer Vision (ICCV)}, pages 558--567, 2021.

\bibitem[Yun et~al.(2020)Yun, Bhojanapalli, Rawat, Reddi, and Kumar]{Yun2020Are}
Chulhee Yun, Srinadh Bhojanapalli, Ankit~Singh Rawat, Sashank Reddi, and Sanjiv Kumar.
\newblock Are transformers universal approximators of sequence-to-sequence functions?
\newblock In \emph{International Conference on Learning Representations}, 2020.

\bibitem[Zeng et~al.(2020)Zeng, Zhou, Srivastava, Kannan, and Prasanna]{zeng2019graphsaint}
Hanqing Zeng, Hongkuan Zhou, Ajitesh Srivastava, Rajgopal Kannan, and Viktor Prasanna.
\newblock Graphsaint: Graph sampling based inductive learning method.
\newblock In \emph{International Conference on Learning Representations}, 2020.

\bibitem[Zhang et~al.(2021)Zhang, Liu, Wang, Lu, and Lee]{zhang2021motif}
Zaixi Zhang, Qi~Liu, Hao Wang, Chengqiang Lu, and Chee-Kong Lee.
\newblock Motif-based graph self-supervised learning for molecular property prediction.
\newblock \emph{Advances in Neural Information Processing Systems}, 34:\penalty0 15870--15882, 2021.

\bibitem[Zhao et~al.(2021)Zhao, Li, Wen, Wang, Liu, Sun, Xie, and Ye]{zhao2021gophormer}
Jianan Zhao, Chaozhuo Li, Qianlong Wen, Yiqi Wang, Yuming Liu, Hao Sun, Xing Xie, and Yanfang Ye.
\newblock Gophormer: Ego-graph transformer for node classification.
\newblock \emph{arXiv preprint arXiv:2110.13094}, 2021.

\bibitem[Zheng et~al.(2024)Zheng, Fu, Maciejewski, and He]{DBLP:conf/TMLR/ZhengFMH24}
Lecheng Zheng, Dongqi Fu, Ross Maciejewski, and Jingrui He.
\newblock Drgnn: Deep residual graph neural network with contrastive learning.
\newblock In \emph{TMLR}, 2024.

\bibitem[Zhu et~al.(2020{\natexlab{a}})Zhu, Yan, Zhao, Heimann, Akoglu, and Koutra]{zhu2020beyond}
Jiong Zhu, Yujun Yan, Lingxiao Zhao, Mark Heimann, Leman Akoglu, and Danai Koutra.
\newblock Beyond homophily in graph neural networks: Current limitations and effective designs.
\newblock \emph{Advances in neural information processing systems}, 33:\penalty0 7793--7804, 2020{\natexlab{a}}.

\bibitem[Zhu et~al.(2021)Zhu, Rossi, Rao, Mai, Lipka, Ahmed, and Koutra]{zhu2021graph}
Jiong Zhu, Ryan~A Rossi, Anup Rao, Tung Mai, Nedim Lipka, Nesreen~K Ahmed, and Danai Koutra.
\newblock Graph neural networks with heterophily.
\newblock In \emph{Proceedings of the AAAI conference on artificial intelligence}, volume~35, pages 11168--11176, 2021.

\bibitem[Zhu et~al.(2020{\natexlab{b}})Zhu, Xu, Yu, Liu, Wu, and Wang]{zhu2020deep}
Yanqiao Zhu, Yichen Xu, Feng Yu, Qiang Liu, Shu Wu, and Liang Wang.
\newblock Deep graph contrastive representation learning.
\newblock \emph{arXiv preprint arXiv:2006.04131}, 2020{\natexlab{b}}.

\bibitem[Zopf(2022)]{zopf20221}
Markus Zopf.
\newblock 1-wl expressiveness is (almost) all you need.
\newblock In \emph{2022 International Joint Conference on Neural Networks (IJCNN)}, pages 1--8, 2022.

\end{thebibliography}

\clearpage
\newpage
\beginappendix

\section{Preliminaries}

\textbf{Graph Attention Networks (GAT)}. The representation of node $i$ in layer $l$ is computed as:
\begin{equation}
    h_i^l=\sigma\left(\sum_{j \in \mathcal{N}_i} \alpha_{ij} \textbf{W} h_j^{(l-1)}\right),  
    \quad
    \alpha_{ij}=\frac{\exp\left(\sigma\left(\textbf{W}_2 \left[\textbf{W}_1h_i^{(l-1)}\Vert\textbf{W}_1h_j^{(l-1)}\right] \right) \right)}
    {\sum\limits_{k \in \mathcal{N}_i} \exp\left(\sigma\left(\textbf{W}_2 \left[\textbf{W}_1h_i^{(l-1)}\Vert\textbf{W}_1h_k^{(l-1)}\right] \right)\right)}
\end{equation}
where $\sigma$ is a non-linearity, and $\alpha_{ij}$ is the normalized attention score between two connected nodes $i$ and $j$.

\textbf{Personalized PageRank (PPR)}. A PPR vector for a node $u$ captures the relative importance of other nodes with respect to node $u$ by exploring the graph structure through iterative random walks:
\begin{equation}
\label{eq: ppr}
    r = \alpha \mathbf{P} r + (1-\alpha) q
\end{equation}
where $\mathbf{P}=\mathbf{D}^{-\frac{1}{2}}\mathbf{A}\mathbf{D}^{-\frac{1}{2}} \in \mathbb{R}^{n \times n}$, $q$ is a stochastic personalized vector, $r$ is the  stationary distribution of random walks, and $\alpha$ is a damping factor.

\section{Model Details}\label{sec:appendix_model_details}

\begin{algorithm}[H]
\caption{Graph Tokenizer}
\label{alg:tokenizer}
  \begin{algorithmic}[1]
    \STATE {\bfseries Input:} Graph $g = \left(\mathcal{V}, \mathcal{E}, \textbf{X} \right)$, Graph Encoder $\text{GNN}_\theta$, Residual Quantizer $\text{RQ}_\Phi$, BGRL Loss $\text{RQ}_\Phi$
    \STATE $\mathbf{H_v}=\text{GNN}_\theta(g)$\hfill\COMMENT{Node Representations}
    \STATE $\mathbf{C}, \mathbf{Z}, \mathbf{T}, \mathcal{L}_{commit} = \text{RQ}_\Phi(\mathbf{H_v})$\hfill\COMMENT{codebooks, quantized representation, discrete tokens}
    \STATE $\mathcal{L}_{dgi}=\text{DGI}(\mathbf{Z})$\hfill\COMMENT{Compute DGI Loss}
    \STATE $\mathcal{L}_{bgrl}=\text{BGRL}(\mathbf{Z})$\hfill\COMMENT{Compute BGRL Loss}
    \STATE $\mathcal{L}_{mae}=\text{MAE}(\mathbf{Z})$\hfill\COMMENT{Compute MAE Loss}
    \STATE $\mathcal{L} = \mathcal{L}_{dgi} + \mathcal{L}_{bgrl} + \mathcal{L}_{mae} + \beta \times \mathcal{L}_{commit}$\hfill\COMMENT{Compute Multi-Task Loss}
    \RETURN $\mathbf{Z}$, $\mathbf{T}$, $\mathcal{L}$
\end{algorithmic}
\end{algorithm}

\begin{algorithm}[H]
\caption{Residual Vector Quantization}
\label{alg:RVQ}
  \begin{algorithmic}[1]
    \STATE {\bfseries Input:} Data $\mathbf{X}$, Number of codebooks $N$, Size of codebook $K$, Dimension of codebooks $d$
    \STATE $\mathcal{L}_\text{commit} = 0$
    \STATE $\mathbf{C} =$ Random($N$, $K$, $d$)   
    \STATE $\mathbf{Z} =$ Zeros($|\mathbf{X}|$, $d$)
    \STATE $\mathbf{T} =$ Zeros($|\mathbf{X}|$, $N$)
    \FOR{$i$ in $|\mathbf{X}|$}
        \STATE $r = \mathbf{X}[i]$
        \FOR{$j=1$ to $N$}
            \STATE $k=\argmin_k \Vert r - \mathbf{C}[j, k] \Vert_2^2$
            \STATE $r = \Vert r-\mathbf{C}[j, k] \Vert_2^2$
            \STATE $\mathbf{T}[i][j]=k$
            \STATE $\mathbf{Z}[i] = \mathbf{Z}[i] + \mathbf{C}[j][k]$
            \STATE $\mathcal{L}_\text{commit} = \mathcal{L}_\text{commit} + \Vert r- \text{sg}[\mathbf{C}[j, k]] \Vert_2^2$
        \ENDFOR
    \ENDFOR
    \RETURN $\mathbf{Z}$, $\mathbf{T}$, $\mathcal{L}/|\mathbf{X}|$
\end{algorithmic}
\end{algorithm}

\section{Datasets} \label{appendix:datasets}
We provide a detailed description of the datasets used in this study. All datasets are publicly available.

\begin{itemize}
\item \textbf{CoraFull}~\citep{bojchevski2017deep}, \textbf{CiteSeer}, and \textbf{Pubmed}~\citep{namata2012query} are citation datasets, where nodes represent documents and edges represent citation links. Labels indicate the paper category.

\item \textbf{Computer} and \textbf{Photo}~\citep{shchur2018pitfalls} are from the Amazon co-purchase graph~\citep{mcauley2015image}, where nodes represent goods and edges indicate that two goods are frequently bought together. Node features are bag-of-words encoded product reviews, and class labels are given by the product category.

\item \textbf{CS} and \textbf{Physics}~\citep{shchur2018pitfalls} are co-authorship graphs based on the Microsoft Academic Graph from the KDD Cup 2016 challenges. Here, nodes are authors connected by an edge if they co-authored a paper; node features represent paper keywords for each author’s papers, and class labels indicate the most active fields of study for each author.

\item \textbf{WikiCS}~\citep{mernyei2020wiki} is derived from Wikipedia, where nodes represent Computer Science articles, and edges are based on hyperlinks. The nodes are classified into 10 classes representing different branches of the field.

\item \textbf{Squirrel} and \textbf{Chameleon}~\citep{rozemberczki2021multi, pei2020geom} are Wikipedia page-page networks, where nodes represent articles from the English Wikipedia, and edges reflect mutual links between them. The nodes were classified into five classes based on their average monthly traffic.

\item \textbf{Amazon-Ratings}~\citep{platonov2023critical} is based on Amazon product co-purchasing data. Nodes represent products (books, music CDs, DVDs, VHS video tapes), and edges connect products that are frequently bought together. The task is to predict the average rating given to a product by reviewers.

\item \textbf{Roman-Empire}~\citep{platonov2023critical} is based on the Roman Empire article from the English Wikipedia. Each node in the graph corresponds to one word (not necessarily unique) in the text, so the number of nodes equals the length of the article. Two words are connected if they follow each other in the text or are linked in the sentence's dependency tree. A node's class represents its syntactic role.

\item \textbf{Minesweeper}~\citep{platonov2023critical} is inspired by the Minesweeper game. The graph consists of regular 100x100 grid, where each node (cell) is connected to eight neighboring nodes (except for nodes at the edge of the grid, which have fewer neighbors). 20\% of the nodes are randomly selected as mines. The task is to predict which nodes are mines. Node features are one-hot-encoded numbers of neighboring mines, however, for 50\% of the nodes, these features are unknown, indicated by a separate binary feature.

\item \textbf{Questions}~\citep{platonov2023critical} is based on data from the question-answering website Yandex Q, where nodes represent users, and edges connect two nodes if one user answered another user’s question during a one-year time interval. The task is to predict which users remained active on the website, forming a binary classification task.

\item \textbf{ogbn-proteins}~\citep{hu2020open} is a protein-protein association network, where nodes represent proteins, and edges indicate biologically meaningful associations between proteins, such as physical interactions, co-expression, or homology. The task is to predict the presence of protein functions in a multi-label binary classification setup.

\item \textbf{ogbn-arxiv}~\citep{hu2020open} is a citation network between all Computer Science (CS) arXiv papers indexed by MAG~\citep{wang2020microsoft}. Each node presents an arXiv paper, and directed edges indicate that one paper cites another. The task is to predict the 40 subject areas of arXiv CS papers, such as cs.AI, cs.LG, and cs.OS.

\item \textbf{ogbn-products}~\citep{hu2020open} is an Amazon product co-purchasing network\footnote{\url{http://manikvarma.org/downloads/XC/XMLRepository.html}} of 2 million products. Edges indicate that products are purchased together. The task is to predict the product category.

\item \textbf{pokec}~\citep{leskovec2016snap, lim2021large} is a social network, where nodes represent users, and edges represent friendships. The task is to predict the gender of users.

\item \textbf{Peptides-Func} is a peptide dataset retrieved from SATPdb \citep{singh2016satpdb} with over 15K peptides. Each node corresponds to a heavy atom, and edges are chemical bonds. The task is to predict 10 peptide functions, forming a multi-label graph classification task.

\item \textbf{Peptides-Struct} consists the same graphs as Peptides-Struct, but with different task. Here the task is to predict aggregated 3D properties (i.e., mass, valence) of the peptides at the graph level.

\item \textbf{COCO-SP} is a node classification dataset based on the MC COCO image dataset \citep{lin2014microsoft}. Each node corresponds to a region of the image belonging to a particular class. These superpixels nodes are extracted with the SLIC algorithm \citep{achanta2012slic}, and two nodes are connected with an edge if the node regions share a
common boundary. The task is to predict the semantic segmentation label for each superpixel node out of 81 classes.

\item \textbf{PCQM-Contact} is a molecule dataset with over 529K molecules \citep{nakata2017pubchemqc}. Atoms are nodes, and chemical bonds are edges. The task is to predict pairs of nodes that will be contacting with each other in the 3D space. 

\end{itemize}

For CoraFull, Pubmed, PubMed, Computer, Photo, CS, and Physics, we follow previous work and use 60\%/20\%/20\% train/valid/test split. For WiKiCS, we follow the official split in \citet{mernyei2020wiki}. For Squirrel, Chameleon, Amazon-Ratings, Roman-Empire, Minesweeper, and Questions, we follow the splits in \citet{platonov2023critical}. For ogbn-proteins, ogbn-arxiv, and ogbn-products, we follow the splits in \citet{hu2020open}. For pokec, we follow the split used in \citet{lim2021large}. For Peptides-Func, Peptides-Struct, COCO-SP, and PCQM-Contact, we follow the split provided in \citet{dwivedi2022long}.

\section{Experimental Setup} \label{appendix:hyperparameters}

\textbf{Software \& Hardware.} GQT is implemented using PyTorch\footnote{\url{https://pytorch.org/}}, PyG\footnote{\url{https://pyg.org/}}, DGL\footnote{\url{https://www.dgl.ai/}}, and the vector-quantize-pytorch package\footnote{\url{https://github.com/lucidrains/vector-quantize-pytorch}}. Most datasets can be accessed through PyG and DGL. All experiments are conducted on a single Nvidia A100 GPU.

\textbf{Hyperparameters \& Experimental Details.} As illustrated in Figure~\ref{fig:overview}, our method consists of two parts: the tokenizer and the Transformer encoder. We provide the hyperparameters and experimental details for each part below.

During the training of the graph tokenizer, we use full-graph training for small- and medium-scale datasets, and apply sampling for large-scale graphs. We consider different sampling methods, including random partitioning, which randomly samples nodes within a graph and returns their induced subgraph; neighbor sampling~\citep{hamilton2017inductive}, GraphSAINT~\citep{zeng2019graphsaint}, and local clustering \citet{hou2023graphmae2}. For the GNN encoder and decoder, we use GCN or GAT as our backbone and tune the number of layers from \{1, 2, 3, 4, 5, 6, 7, 8, 9, 10\} and hidden dimensions from \{128, 256, 512, 1024\}. For the quantizer, we use residual-VQ (RVQ)~\citep{lee2022autoregressive} and tune the number of codebooks from \{1, 2, 3, 6, 9\} and the codebook size from \{128, 256, 512, 1024, 2048, 4096\}. We set the code dimension to be equal to the hidden dimension of the GNN encoder.

During the training of the Transformer, we use KNN to add semantic edges and tune the number of semantic neighbors from \{0, 5, 10, 15, 20\}. Then, we use PPR to generate a sequence of nodes for each target node. We tune the number of PPR neighbors from \{0, 5, 10, 20, 30, 50\}. For the Transformer model, we use the TransformerEncoder module in PyTorch as our backbone, and tune the number of layers from\{1, 2, 3, 4, 5, 6\}, the number of heads from \{4, 8\}, and the feedforward dimension from \{512, 1024, 2048\}. Note that for some small- and medium-scale datasets, we do not need PPR-based sequences, instead, we can directly serialize all nodes within the graph as in \citet{rampavsek2022recipe}. 

\begin{table}[ht]
    \setlength{\tabcolsep}{3pt}
    \centering
    \caption{Selected hyperparameters for each dataset.}
    \resizebox{\textwidth}{!}{
    \begin{tabular}{lccccccccc}\toprule
        &\multicolumn{2}{c}{\textbf{GNN Encoder}} &\multicolumn{2}{c}{\textbf{Quantizer}} &\multicolumn{5}{c}{\textbf{Transformer}} \\\midrule
        &\# layers &\# Hidden dim &\# Codebooks &Codebook size &KNN &PPR &\# Layers &\# Heads &\# FFN dim \\\midrule
        CoraFull &2 &256 &3 &128 &0 &15 &2 &4 &512 \\
        CiteSeer &2 &256 &3 &128 &5 &15 &2 &4 &512 \\
        PubMed &2 &256 &3 &256 &0 &15 &2 &4 &512 \\
        Computer &2 &256 &3 &128 &5 &30 &2 &4 &512 \\
        Photo &3 &512 &3 &128 &5 &30 &2 &4 &1024 \\
        CS &2 &512 &3 &128 &5 &20 &2 &4 &1024 \\
        Physics &2 &256 &3 &256 &5 &30 &2 &4 &512 \\
        WikiCS &2 &256 &3 &128 &5 &30 &2 &4 &512 \\
        Squirrel &3 &256 &3 &128 &5 &30 &2 &4 &512 \\
        Chameleon &3 &256 &3 &128 &5 &30 &2 &4 &512 \\
        Amazon-Ratings &4 &512 &3 &128 &5 &20 &2 &4 &1024 \\
        Roman-Empire &6 &256 &3 &256 &10 &15 &3 &4 &512 \\
        Minesweeper &6 &128 &3 &128 &10 &15 &2 &4 &512 \\
        Questions &3 &256 &3 &512 &10 &15 &2 &4 &512 \\
        ogbn-proteins &6 &256 &3 &512 &0 &50 &3 &4 &512 \\
        ogbn-arxiv &4 &512 &3 &512 &5 &30 &2 &4 &1024 \\
        ogbn-products &4 &1024 &3 &4096 &5 &30 &2 &8 &2048 \\
        pokec &6 &256 &3 &512 &0 &50 &3 &4 &512 \\
        Peptides-Func &4 &128 &3 &128 &0 &0 &2 &4 &512 \\
        Peptides-Struct &4 &128 &3 &128 &0 &0 &2 &4 &512 \\
        COCO-SP &4 &128 &3 &128 &0 &0 &2 &4 &512 \\
        PCQM-Contact &4 &128 &3 &128 &0 &0 &2 &4 &512 \\
        \bottomrule
    \end{tabular}}
    \label{tab:selected_hyperparameters}
\end{table}

\section{Additional Results} \label{appendix:additional_results}
\subsection{Further Ablation Study} \label{appendix:ablation}
We also provide an ablation study on one of the heterophilic datasets. The results shown in Table~\ref{tab:appendix_ablation} suggest that introducing semantic edges and structural gating mechanisms specifically benefits the heterophilic setting.

\begin{table}[h]
\caption{\small Ablation study on effect of proposed components on the Minesweeper dataset.}
\vspace{-0.1in}
\label{tab:appendix_ablation}
\begin{center}
\resizebox{\textwidth}{!}{
\begin{tabular}{lccc ccc cc c c}\toprule
 & \multicolumn{3}{c}{\textbf{Graph Tokenizer}} & \multicolumn{3}{c}{\textbf{Token Modulation}} & \multicolumn{2}{c}{\textbf{Augmentation}} &\textbf{Model} & \textbf{Performance}\\
 \cmidrule{2-9}\cmidrule{11-11}
 & RVQ	& GMAE2	& DGI & Codebook   & Positional & Structural & Semantic & PPR      & & ROC-AUC$\uparrow$ \\
 &     &	    &	  &	Embeddings & Encoding	& Gating	 & Edges	& Sequence & & \\
\midrule
(1)  & \cmark &\cmark &\cmark & \cmark & & & & & Linear & 90.24$\pm$0.49 \\
(2)  &  &  & & &\cmark & & &\cmark & Transformer & 90.52$\pm$0.39  \\

\midrule

(3)  & &\cmark &\cmark &\cmark & \cmark & \cmark&\cmark & \cmark & Transformer & 95.27$\pm$0.46  \\
(4)  & \cmark & &\cmark &\cmark & \cmark & \cmark&\cmark & \cmark & Transformer & 92.91$\pm$0.55  \\
(5)  & \cmark & \cmark & &\cmark & \cmark & \cmark &\cmark & \cmark & Transformer & 93.82$\pm$0.46  \\
\midrule

(6)  & \cmark &\cmark &\cmark &  &\cmark & \cmark& \cmark &\cmark & Transformer & 93.24$\pm$0.36 \\
(7)  & \cmark  &\cmark &\cmark &\cmark &  & \cmark &\cmark & \cmark & Transformer & 94.82$\pm$0.41 \\
(8)  & \cmark & \cmark & \cmark &\cmark & \cmark & &\cmark & \cmark & Transformer & 93.97$\pm$0.58 \\
\midrule

(9)  & \cmark &\cmark &\cmark &\cmark &\cmark &\cmark & &\cmark & Transformer &92.83$\pm$0.35 \\
\midrule
(10) & \cmark & \cmark & \cmark &\cmark & \cmark & \cmark &\cmark & \cmark & Transformer & 95.28$\pm$0.44\\
\bottomrule
\vspace{-0.3in}
\end{tabular}}
\end{center}
\end{table}

\newpage

\subsection{Generalization Analysis} \label{appendix:generalization}
\begin{wraptable}{r}{0.4\textwidth}
    \vspace{-0.15in}
    \setlength{\tabcolsep}{3pt}
    \centering
    \caption{Comparison between mean GQT and RQ-VAE performance over five runs.}
    \resizebox{0.4\textwidth}{!}{
    \begin{tabular}{lcc}\toprule
    ~ & ogbn-arxiv & Minesweeper \\\midrule
    RQ-VAE &66.05$\pm$0.48 &89.69$\pm$0.35 \\
    GQT (ours) & 73.14$\pm$0.16 & 95.28$\pm$0.44 \\
    \bottomrule
    \end{tabular}
    }
    \label{tab:generalization}
\end{wraptable}
To measure improved generalization, we follow the common practice of treating downstream predictive performance as a proxy for generalization. As shown in Table~\ref{tab:ablation} and Table~\ref{tab:appendix_ablation}, every component of the tokenizer, including both SSL objectives and the quantization layer, contributes to the downstream predictive performance, thereby improving the model’s generalizability. Furthermore, to evaluate the contribution of multi-task SSL objectives to downstream performance, we compare our results with those of a tokenizer trained using the RQ-VAE~\citep{lee2022autoregressive} design, which employs a reconstruction objective. The results presented below indicate that using multi-task SSL objectives significantly improves downstream predictive performance, which is strongly correlated with the method's generalization. 

\subsection{Efficiency Analysis}
\begin{wraptable}{r}{0.6\textwidth}
    \vspace{-0.15in}
    \setlength{\tabcolsep}{3pt}
    \centering
    \caption{Memory and run time during inference.}
    \resizebox{0.6\textwidth}{!}{
    \begin{tabular}{lcccc}\toprule
    Attack & \multicolumn{2}{c}{GPU Memory} & \multicolumn{2}{c}{Full Inference Time}\\\midrule
    & ogbn-arxiv & Minesweeper & ogbn-arxiv & Minesweeper \\\midrule
    GAT &   2715MB & 2108M &5s & 1s \\
    GQT (ours) & 1324MB &1037MB & 4s & 1s \\
    \bottomrule
    \end{tabular}}
    \label{tab:efficiency}
\end{wraptable}
As mentioned in Section~\ref{sec:main_results}, using discrete tokens instead of node features results in significant memory reduction. For instance, on the ogbn-products dataset with 2,449,029 nodes and 100-dimensional node features, GQT requires only 3 codebooks of size 4096, resulting in a remarkable 30-fold reduction in memory usage. This memory reduction occurs after training the tokenizer. Since the encoder of the tokenizer is a GNN that processes the graph with original node features, its memory footprint is comparable to that of any arbitrary GNN. However, because the Transformer encoder only consumes discrete tokens, which are significantly fewer than the total number of nodes, we achieve a substantial reduction in memory footprint. As an additional experiment, we compare the inference time and memory usage between our Transformer encoder and a Graph Attention Network (GAT) when performing inference on all graph nodes. The results shown in Table \ref{tab:efficiency} show that while our Transformer is on par with a sparse implementation of GAT in terms of inference time, it requires half the GPU memory.
\end{document}